\documentclass[10pt,twocolumn,letterpaper]{article}

\usepackage{authblk}
\usepackage{wacv}              

\usepackage{graphicx}
\usepackage{amsmath}
\usepackage{amssymb}
\usepackage{booktabs}

\usepackage[pagebackref,breaklinks,colorlinks]{hyperref}

\usepackage[capitalize]{cleveref}
\crefname{section}{Sec.}{Secs.}
\Crefname{section}{Section}{Sections}
\Crefname{table}{Table}{Tables}
\crefname{table}{Tab.}{Tabs.}

\usepackage{multirow}
\usepackage[table]{xcolor} 

\def\wacvPaperID{864} 
\def\confName{WACV}
\def\confYear{2025}

\begin{document}

\title{Pairwise Comparisons Are All You Need}





\author{
    Nicolas Chahine$^{1,2}$\hspace{2em}Sira Ferradans$^{1}$\hspace{2em}Jean Ponce$^{2,3}$\\
    
    \normalsize{$^{1}$DXOMARK} \hspace{2em}
    \normalsize{$^{2}$Département d'informatique de l'Ecole normale supérieure (ENS-PSL, CNRS, Inria)}\\
    \normalsize{$^{3}$Institute of Mathematical Sciences and Center for Data Science, New York University}
}


\maketitle

\begin{abstract}
Blind image quality assessment (BIQA) approaches, while promising for automating image quality evaluation, often fall short in real-world scenarios due to their reliance on a generic quality standard applied uniformly across diverse images. This one-size-fits-all approach overlooks the crucial perceptual relationship between image content and quality, leading to a 'domain shift' challenge where a single quality metric inadequately represents various content types. Furthermore, BIQA techniques typically overlook the inherent differences in the human visual system among different observers. In response to these challenges, this paper introduces PICNIQ, a pairwise comparison framework designed to bypass the limitations of conventional BIQA by emphasizing relative, rather than absolute, quality assessment. PICNIQ is specifically designed to estimate the preference likelihood of quality between image pairs. By employing psychometric scaling algorithms, PICNIQ transforms pairwise comparisons into just-objectionable-difference (JOD) quality scores, offering a granular and interpretable measure of image quality. The proposed framework implements a deep learning architecture in combination with a specialized loss function, and a training strategy optimized for sparse pairwise comparison settings.  We conduct our research using comparison matrices from the PIQ23 dataset, which are published in this paper. Our extensive experimental analysis showcases PICNIQ's broad applicability and competitive performance, highlighting its potential to set new standards in the field of BIQA.
\end{abstract}

\section{Introduction}
\label{sec:intro}
Smartphones are closing the gap with professional photography by integrating modern image enhancement technology through sophisticated non-linear processes \cite{van2019edge, morikawa2021image, li2023diffusion}. Consequently, smartphone cameras produce images with new realistic distortions that are difficult to model, unlike classical linear imaging systems. This makes traditional image quality assessment (IQA) methods \cite{loebich2007digital, gousseau2007modeling, cao2009measuring, nicolas2021portrait}, that model digital cameras as linear systems, unreliable for camera tuning \cite{fang2020perceptual}. Therefore, in addition to objective IQA measurements, smartphone manufacturers conduct rigorous tuning procedures to optimize the quality of their camera systems \cite{dxomarklabs}. The evaluation process, usually referred to as perceptual quality assessment, consists of shooting and evaluating thousands of use cases, which can be costly, time-consuming, and challenging to reproduce. 

As an alternative to perceptual evaluation, blind IQA (BIQA) methods try to mimic human perception \cite{moorthy2011blind, saad2012blind, mittal2012no, ye2012unsupervised, ghadiyaram2017perceptual, xue2013learning, mittal2012making, zhang2015feature}, promising to deliver universal image quality metrics. 
Learning-based BIQA methods \cite{yang2023deep}, 
in particular, have shown marked improvements over their classical counterparts, due to their ability to extract perceptual image quality information. However, existing BIQA solutions often use a one-size-fits-all approach for quality by applying the same concepts to different content, without considering the impact of scene-specific semantics on the quality \cite{su2020blindly, zhang2021uncertainty, chahine2023image}. Given that image quality varies under different conditions, numerous studies have underscored the significance of integrating semantic information into IQA, tackling a challenge known as domain shift \cite{sun2021blind, zerman2017extensive, zhang2020learning, zhang2021uncertainty, chahine2023image, fang2020perceptual, feng2023learning, huang2020multi, su2020blindly, chahine2024generalized}. Moreover, some approaches attend to model the uncertainty of image quality predictions \cite{talebi2018nima, liu2019comprehensive, wu2017blind, zhang2021uncertainty, ma2017dipiq, ma2019blind}, due to the intrinsic inconsistency of the human's perception of quality.

An extra challenge comes from the available IQA datasets. 
Most datasets \cite{sheikh2006statistical, larson2010most, ponomarenko2015image, ponomarenko2009tid2008, zhang2018unreasonable, virtanen2014cid2013, ghadiyaram2015massive, zhu2020multiple, hosu2020koniq, ying2020patches, fang2020perceptual} are annotated using rating-based procedures, such as mean opinion scores (MOS), which are commonly known to be noisy and non-deterministic \cite{mantiuk2012comparison}.
There is thus a clear need for a quantitative and formal framework to evaluate and compare perceptual judgments objectively. A recent dataset annotated by image quality experts, PIQ23 \cite{chahine2023image}, promises high precision by adopting content-dependent pairwise comparisons to annotate image quality, a method known for enhancing consistency in IQA experiments \cite{mantiuk2012comparison, perez2019pairwise}. 

In tackling the limitations of current BIQA methodologies, such as combining IQA knowledge, limitations on generalization, domain shift, and uncertainty, we present PICNIQ, a BIQA framework purely based on pairwise comparisons. Specifically, instead of direct quality score predictions, our model adopts a pairwise ranking approach, simplifying the task to predicting the preference likelihood between a pair of images, while retaining the ability to interpret quality differences. Through psychometric scaling algorithms applied to the pairwise comparisons, we transform the rankings into a granular quality scale, facilitating a robust and interpretative assessment of image quality. Our findings highlight the potential of this approach in addressing the identified challenges in IQA, contributing towards the development of more reliable and generalized image quality evaluation models. We summarize our contributions as follows:

\begin{itemize}
\item We introduce PICNIQ, a novel BIQA model that estimates the pairwise preference likelihood of quality. By employing probabilistic ranking, PICNIQ ensures interpretable quality predictions, accurately estimates the uncertainty of quality differences, and effectively addresses limitations related to domain shift and cross-content generalization.

\item Our pairwise comparison framework incorporates a unique training strategy, comparison loss, and inference approach. We demonstrate the necessity of using a weighted binary cross-entropy loss to train in sparse comparison settings. Additionally, we define an inference pipeline that integrates PICNIQ to derive quality scores from pairwise preferences using psychometric scaling and active sampling algorithms.

\item We explore the sparse pairwise comparison matrices in the PIQ23 dataset. Our approach exploits the granular insights of image quality present in PIQ23, compared to the single image quality scores in other datasets.

\item We conduct a comprehensive evaluation of PICNIQ, demonstrating its competitive performance against existing benchmarks. Our analysis highlights PICNIQ's ability to generalize to new, unseen conditions while overcoming previous challenges in BIQA.

\end{itemize}

\section{Related Work}
\label{sec:related_work}

IQA methods fall into two main categories. Full-reference IQA (FR-IQA) techniques rely on evaluating images based on a pristine reference image, while no-reference (NR-IQA) methods try to capture image quality without any reference. Blind image quality assessment (BIQA) is a part of NR-IQA where neither the reference image nor the distortion space is known. This section reviews both the key datasets and algorithms of BIQA. We also review approaches that aim to quantify the uncertainty in image quality as well as solve the domain shift problem.

\subsection{BIQA datasets}
Traditional datasets such as LIVE \cite{sheikh2006statistical}, CSIQ \cite{larson2010most}, TID \cite{ponomarenko2009tid2008, ponomarenko2015image}, and BAPPS \cite{zhang2018unreasonable} provide reference images with synthetic distortions. Recent datasets like CLIVE \cite{ghadiyaram2015massive}, KonIQ10k \cite{hosu2020koniq}, and PaQ-2-PiQ \cite{ying2020patches} include real-world images, better reflecting realistic image distortions. Other datasets, such as CID2013 \cite{virtanen2014cid2013}, SCPQD2020 \cite{zhu2020multiple} and SPAQ \cite{fang2020perceptual} expand the spectrum of realistic IQA datasets by annotating smartphone images in controlled settings. Despite their contributions, their reliance on MOS often overlooks the content's impact on quality perception and introduces inconsistencies. Chahine \etal's PIQ23 \cite{chahine2023image}, focusing on portrait quality assessment with pairwise comparison annotations, addresses these limitations. Its structured approach and precise annotations make it central to our study, providing valuable comparison matrices for further research.

\subsection{BIQA methods}
BIQA algorithms can be grouped into regression-based and ranking-based methods. While both aim to predict quality scores, ranking-based approaches also consider the model's ability to accurately rank images before assigning scores.

\subsubsection{Regression-based BIQA}
Regression-based methods have evolved from classical techniques using hand-crafted features \cite{moorthy2011blind, saad2012blind, mittal2012no, ye2012unsupervised, ghadiyaram2017perceptual} to deep learning models employing CNNs \cite{kang2014convolutional, kim2016fully, bosse2017deep} and transformers \cite{you2021transformer, ke2021musiq, golestaneh2022no, yang2022maniqa, qin2023data}. Self-supervised learning methods based on contrastive learning \cite{babu2023no, zhao2023quality, saha2023re}, have recently been explored for training universal BIQA models.

\subsubsection{Ranking-based BIQA}
Ranking-based methods treat BIQA as a learning-to-rank problem, utilizing distortion levels, FR-IQA models, and human judgments for relative rankings. Gao \etal \cite{gao2015learning} introduced a multi-kernel learning approach to predict human pairwise preferences for synthetically distorted images and proposed an inference strategy based on a linear combination of multiple pairwise comparisons. The dataset used in this work is not available online. Early deep learning approaches like  RankIQA \cite{liu2017rankiqa} and DBCNN \cite{zhang2018blind} employ Siamese networks pre-trained on synthetically distorted images. Ma \etal's dipIQ \cite{ma2017dipiq} represents an early self-supervised strategy, training on quality-discriminable image pairs (DIPs) gathered via multiple FR-IQA measures. The same authors have also trained another probabilistic CNN model \cite{ma2019blind} on DIPs by treating each FR-IQA model as a separate annotator. Shi \etal \cite{shi2020pairwise} combine objective IQA features for binary classification-based preference prediction. Xu \etal \cite{xu2023blind} introduce a perception-based method for generating pairwise rankings from synthetic distortions, employing an eigenvalue decomposition (EVD) for unsupervised quality score inference. UNIQUE \cite{zhang2021uncertainty} takes an uncertainty-aware cross-domain approach to BIQA, utilizing a Gaussian prior for quality distribution and training on image pairs from multiple datasets. We aim to extend this approach to train on sparse pairwise comparison data using a model-free uncertainty approach. This makes our model more adapted to realistic incomplete pairwise comparison data, where the number of comparisons is usually less than enough to assume a Gaussian prior, therefore optimizing the costs of pairwise experiments. It also allows us to avoid the domain shift in cross-content training and generalization, where the quality scores do not reflect similar quality.

\subsection{Uncertainty-aware BIQA}
Acknowledging the inherent noise and uncertainty in human-based image quality judgments, several studies have sought to integrate these aspects into BIQA metrics. Kendall \etal \cite{kendall2017uncertainties} and Duanmu \etal \cite{duanmu2021quantifying} stress the importance of incorporating probabilistic models to enhance decision-making under uncertainty in computer vision and IQA. Techniques like PQR \cite{zeng2017probabilistic}, NIMA \cite{talebi2018nima} and Liu \etal's work \cite{liu2019comprehensive} employ a label distribution learning approach. Wu \etal \cite{wu2017blind} proposed the LOCRUE, which combines natural scene statistics (NSS) with a Gaussian process to predict image quality. Besides, multiple ranking-based methods, previously mentioned in this paper \cite{zhang2021uncertainty, ma2017dipiq, ma2019blind}, incorporate a probabilistic learning approach, based on the Gaussian distribution assumptions of the Thurstone case V observer model \cite{thurstone1994law, davidson1976bibliography}. These methods are limited due to their reliance on specific data structures, strong prior assumptions, or inability to account for cross-domain IQA. 
We propose a method that accounts for all the previous problems using a model-free pairwise comparison framework.

\subsection{Domain shift challenge in BIQA}
\label{subsub:domainshift}
IQA datasets often feature diverse annotation strategies and image content, leading to relative and independent quality scales. This diversity presents a `domain shift' challenge, complicating cross-content learning and generalization \cite{zerman2017extensive, zhang2020learning, zhang2021uncertainty, sun2021blind}. Multitask learning has been proposed as a solution \cite{huang2020multi, fang2020perceptual, zerman2017extensive, sun2021blind}, but these efforts often fail to explicitly separate semantic information from quality. Su \etal's HyperIQA \cite{su2020blindly} offer a novel approach by using a self-adaptive hypernetwork to implicitly adjust quality predictions based on semantic information. 
Chahine \etal \cite{chahine2023image, chahine2024generalized} extend the HyperIQA architecture for better semantic understanding and generalization by forcing the model to predict the category of each image. Our work addresses the issue of domain shift intrinsically by using a generic pairwise preference framework that can easily be expanded to multiple sources.

\section{PICNIQ}
\label{sec:PICNIQ}

\subsection{Motivation}
\label{sub:motivation}
Blind image quality assessment (BIQA) confronts significant challenges undermining its efficacy in camera evaluation. 
Xu \etal \cite{xu2023blind} categorize these challenges based on the content diversity, the stochastic nature of annotations, and the arbitrary sampling of image distortions. 

Both FR-IQA and NR-IQA methods fail to address the demands for high-precision digital camera evaluation. While the first struggle with the absence of pristine reference images, as a result of the complex non-linearities in modern digital cameras, NR-IQA methods often fail to achieve the precision necessary to be considered reliable IQA metrics. On the one hand, the common reliance on mean opinion scores (MOS) for deriving quality metrics is flawed, due to the subjective and noisy nature of these scores. This subjectivity, alongside factors like annotation guidelines, data collection procedures, visual conditions, and the diverse backgrounds of annotators, complicates the effective handling of domain shifts through direct score prediction methods \cite{wang2021active, zhang2021uncertainty}. 
On the other hand, pairwise comparison models, despite their potential, encounter their own set of obstacles. For instance, their Gaussian assumptions might not hold well in sparse comparison settings. Additionally, they rely on quality scores to derive the comparison probability, which inherits the limitations of MOS in domain shift. Adding to that the random pair selection process and the quality predictions can be significantly impacted. This happens because the image quality is derived from pairwise comparison graphs rather than single comparisons, and there is a limitation on the quality difference allowed to accurately derive comparison probabilities \cite{perez2017practical, perez2019pairwise}. Consequently, most BIQA methods do not adequately tackle the challenges posed above, such as the limited size of IQA datasets, the pairwise comparison constraints, and the domain shifts related to BIQA \cite{xu2023blind}.

To tackle these challenges, we introduce a pairwise comparison framework designed to capture relative quality preferences among image pairs.
Our approach incorporates model-free uncertainty through empirical Bayes principles and leverages pairwise comparison graphs to derive image quality scores. Our strategy marks a notable departure from conventional methodologies and bridges the gap between FR-IQA and NR-IQA methods, promising a more precise and scalable solution for BIQA.

\subsection{Problem formulation}
\label{sub:pb_formulation}
Given a set of images $\{I_1, I_2, \ldots, I_n\}$, we construct a zero-diagonal pairwise comparison matrix $C$, where each element $c_{ij}$ represents the empirical win count of image $I_i$ over $I_j$ according to image quality preference. We construct a model $M$, parameterized by $\theta$, which takes as input a pair of images $(I_i, I_j)$ and outputs the likelihood $M_\theta(I_i, I_j)$ of $I_i$ being of higher quality than $I_j$. We assume $M_\theta$ guarantees probabilistic symmetry, hence:
\begin{equation}
M_\theta(I_j, I_i) = 1 - M_\theta(I_i, I_j).
\label{eq:symmetry}
\end{equation}

In contrast to previous pairwise approaches in BIQA, we do not make any assumptions regarding the prior image quality distribution. Instead, we allow the posterior distributions to be predominantly informed by the available data. More precisely, previous methods such as UNIQUE \cite{zhang2021uncertainty} predict a quality score per image and then try to compute the likelihood based on a Gaussian prior (specifically, Thurstone case V) of the quality distribution. We argue that this is only true when we have a large number of annotators and comparisons in a controlled annotation environment. Moreover, computing probabilities from scores and vice versa can only work for small quality differences, since the normal cumulative distribution function tends towards infinity when the probability tends to 1 \cite{perez2019pairwise}. Finally, the quality of an image cannot be determined solely by comparing it to another image, but it should be evaluated in the context of a broader comparison graph of a larger image set \cite{chahine2023image}. Since we do not assume any prior on the quality distribution, we directly predict the likelihood that one image is better than the other, supposing the model will learn the intrinsic quality distributions from the pairwise training data. Our approach aligns with the empirical Bayes methodology, where priors are broad or estimated from the data, ensuring that the posterior inference is primarily driven by the empirical data available.

\subsection{Loss function}
\label{sub:Loss}
Our pairwise training approach adopts a maximum likelihood estimation (MLE) framework, akin to the methodology proposed by Perez \etal (2019)\cite{perez2019pairwise}. MLE looks for the model parameters that maximize the probability of observing our dataset $\Omega$. We define the true probability of preferring image $I_i$ over $I_j$ as $P_{ij}$. Given $c_{ij}$ occurrences of $I_i$ being preferred in $n_{ij}$ comparisons, Perez \etal propose that the likelihood follows a binomial distribution:
\begin{equation}
L(P_{ij} | c_{ij}, n_{ij}) = \binom{n_{ij}}{c_{ij}} P_{ij}^{c_{ij}} (1 - P_{ij})^{n_{ij} - c_{ij}},
\end{equation}
where $\binom{n_{ij}}{c_{ij}}$ is the binomial coefficient. Given that the binomial coefficient is a multiplicative factor, independent of the model parameters $\theta$, we omit this term in the loss function. 

We define the model's estimated probability that $I_i$ is preferred over $I_j$ as $P^{\theta}{ij} = M_\theta(I_i, I_j)$, and our objective is to minimize the negative log-likelihood of observing the given preferences across all compared pairs in the dataset $\Omega$:

\begin{align}
\theta^* \longleftarrow &\arg \min_{\theta} -\frac{1}{N} \log \left( \prod_{i,j \in \Omega} L(P^{\theta}_{ij} | c_{ij}, n_{ij}) \right) \nonumber \\
& = -\frac{1}{N} \sum_{i=1}^{n} \sum_{j=1, j\neq i}^{n} \Bigg[ c_{ij} \log \left( M_\theta(I_i, I_j) \right) \nonumber \\
& \quad + (n_{ij} - c_{ij}) \log \left(1 - M_\theta(I_i, I_j) \right) \Bigg],
\end{align}

where $N$ is a normalization factor representing the total number of comparisons, ensuring the loss is averaged over all pairs.
By defining the empirical probability of preference of $I_i$ over $I_j$ as $p_{ij} = \frac{c_{ij}}{n_{ij}}$, we can reformulate the empirical risk $l_\theta$ as follows:

\begin{align}
l_\theta = & -\frac{1}{N} \sum_{i=1}^{n} \sum_{j=1, j\neq i}^{n} n_{ij} \Bigg[ p_{ij} \log M_\theta(I_i, I_j) \nonumber \\
& \quad + (1 - p_{ij}) \log (1 - M_\theta(I_i, I_j)) \Bigg],
\end{align}

using a weighted binary cross-entropy (BCE) loss, where $n_{ij}$ acts as a weight accounting for the varied number of comparisons between different pairs. This is a critical distinction from the standard BCE loss, which assumes equal weights (i.e., $n_{ij}=\text{const}$) for all comparison instances. The weighted BCE approach is essential for handling datasets in imbalanced or sparse comparison settings, which is not often considered in other BIQA methods. 

\begin{figure}[t!]
    \centering
    {\includegraphics[width=0.99\columnwidth]{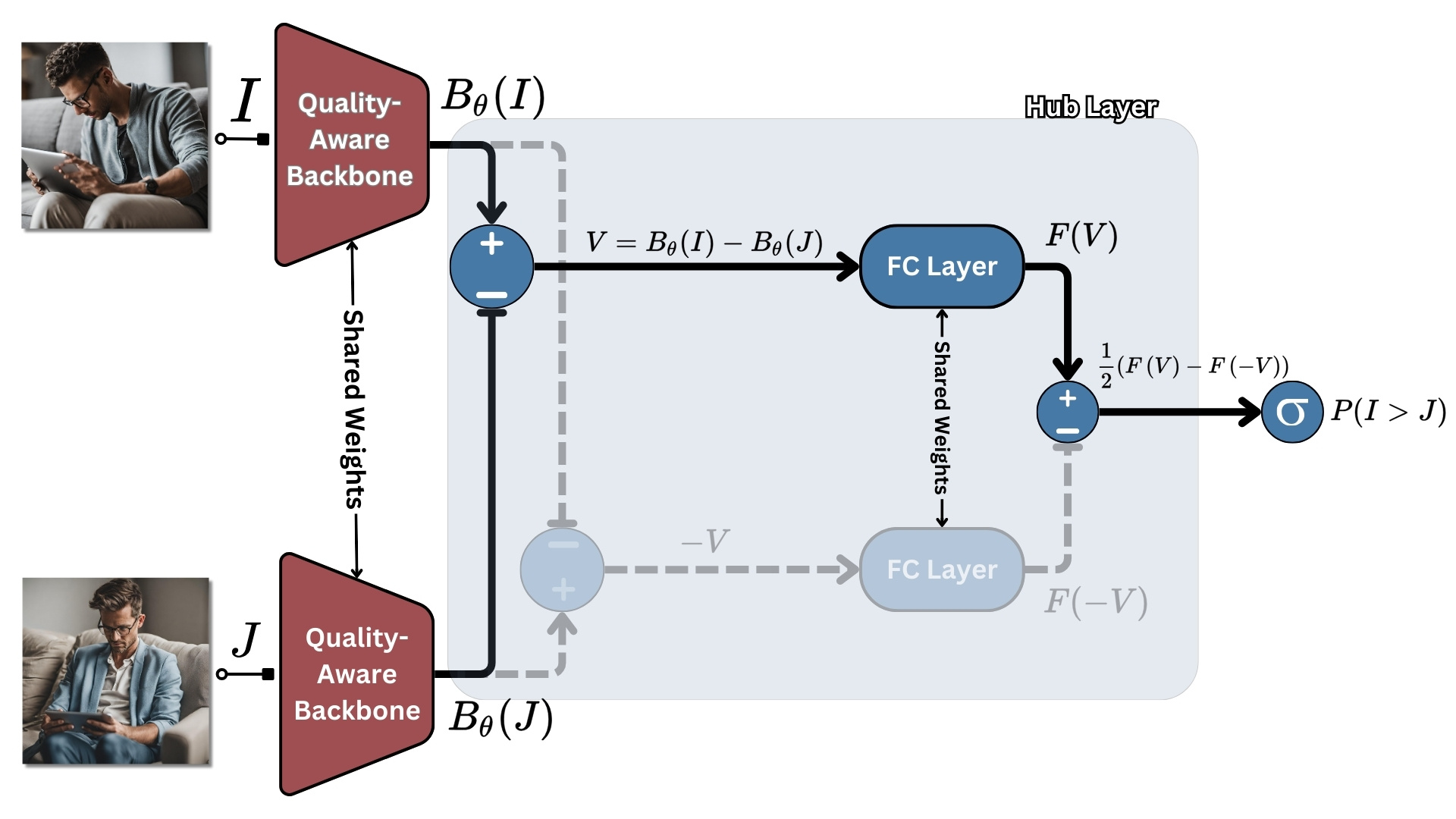}}
    \caption{\footnotesize{The PICNIQ architecture for image quality comparison. A Siamese quality-aware backbone is used to process a pair of images and extract relevant image quality features ($B_\theta$). The difference between the extracted features, $V$, is computed and then fed into a fully connected (FC) layer within the hub layer. The hub layer processes both $V$ and its negation and outputs their difference to ensure probabilistic symmetry in the comparison. The predicted probability $P(I>J)$, is obtained by passing the output of the hub layer through a sigmoid function.}}
    \label{fig:PICNIQ}
\end{figure}

\subsection{Model specifications}
\label{sub:model}
We structure our model to capture quality differences between image pairs, simulating human perceptual comparisons. The proposed model is shown in \Cref{fig:PICNIQ}. First, given two input images $I$ and $J$, we extract their image quality features using a Siamese quality-aware CNN backbone $B_\theta$. We then feed the difference feature vector $V = B_\theta(I) - B_\theta(J)$ to a fully connected layer $H_\theta$, followed by a $\texttt{sigmoid}$ $\sigma$, which gives the likelihood that $I$ has a better quality than $J$, $M_\theta(I, J) = \sigma \left( H_\theta(V) \right)$. We ensure the probabilistic symmetry described in \cref{eq:symmetry}, using a hub layer approach inspired by Mattheakis \etal \cite{mattheakis2019physical}. The hub layer $H_\theta(V) = \frac{1}{2} \left( F_\theta(V) - F_\theta(-V) \right)$, where $F_\theta$ is a fully connected layer, leverages odd properties which when combined with the $\texttt{sigmoid}$'s property $\sigma(H_\theta(V)) + \sigma(-H_\theta(V)) = 1$,  achieves $M_\theta(I, J) = 1 - M_\theta(J, I)$.

In this work, we employ DBCNN's \cite{zhang2018blind} deep dual bilinear architecture as a quality-aware backbone, combined with a simple fully connected hub layer. Since our training data is limited to PIQ23, we do not explore deeper architectures. However, the PICNIQ framework can integrate any quality-aware backbone. As shown by our experiments in \cref{sub:results}, with a simple VGG-16 backbone, PICNIQ achieves competitive generalization performance.

\subsection{Pairwise comparison data}
\label{sub:dataset}

\begin{figure}[!h] 
\setlength{\tabcolsep}{1pt}
    \centering
    \begin{tabular}{cc}
        \includegraphics[width=0.49\columnwidth]{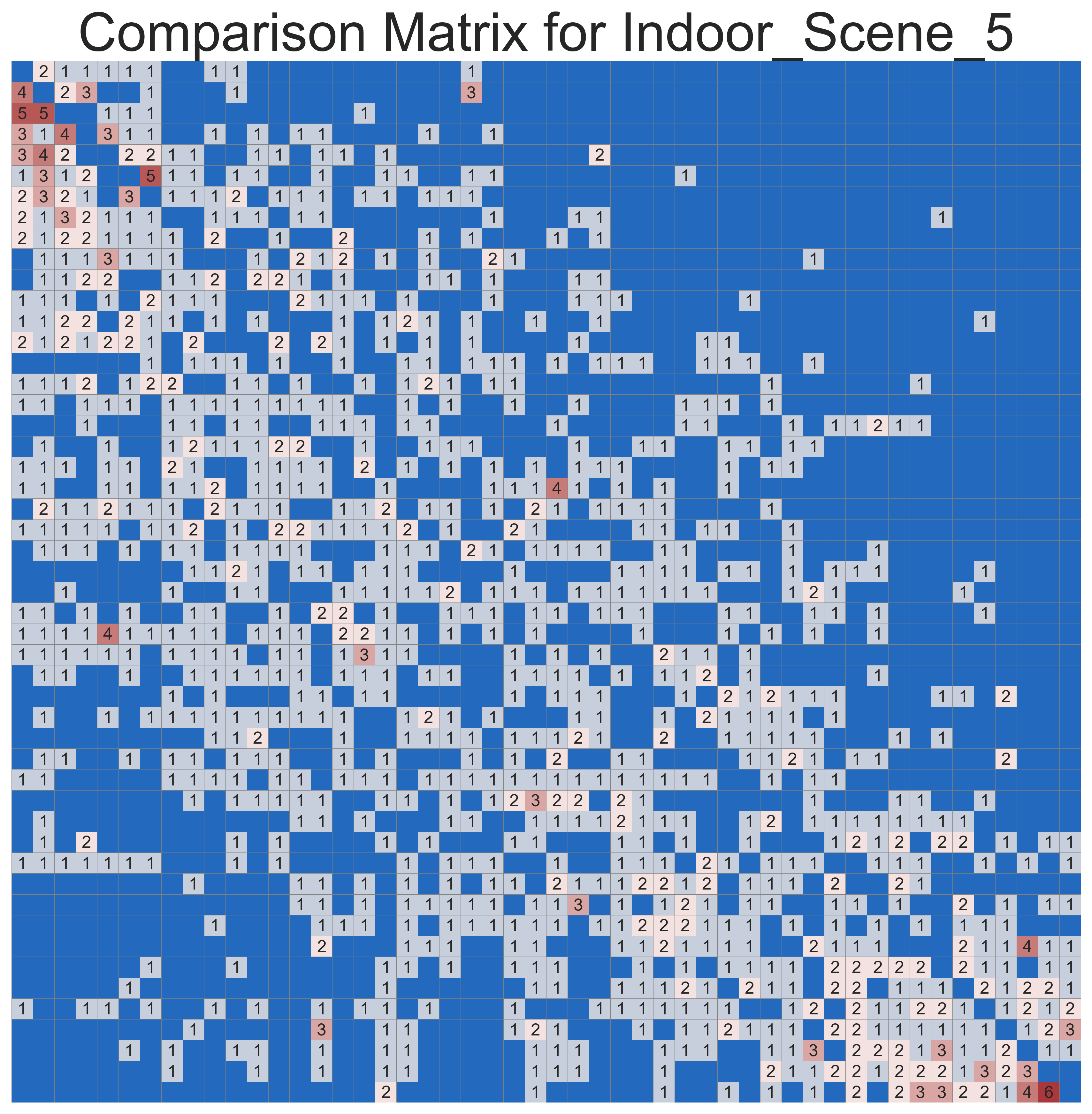}
        &

        \includegraphics[width=0.49\columnwidth]{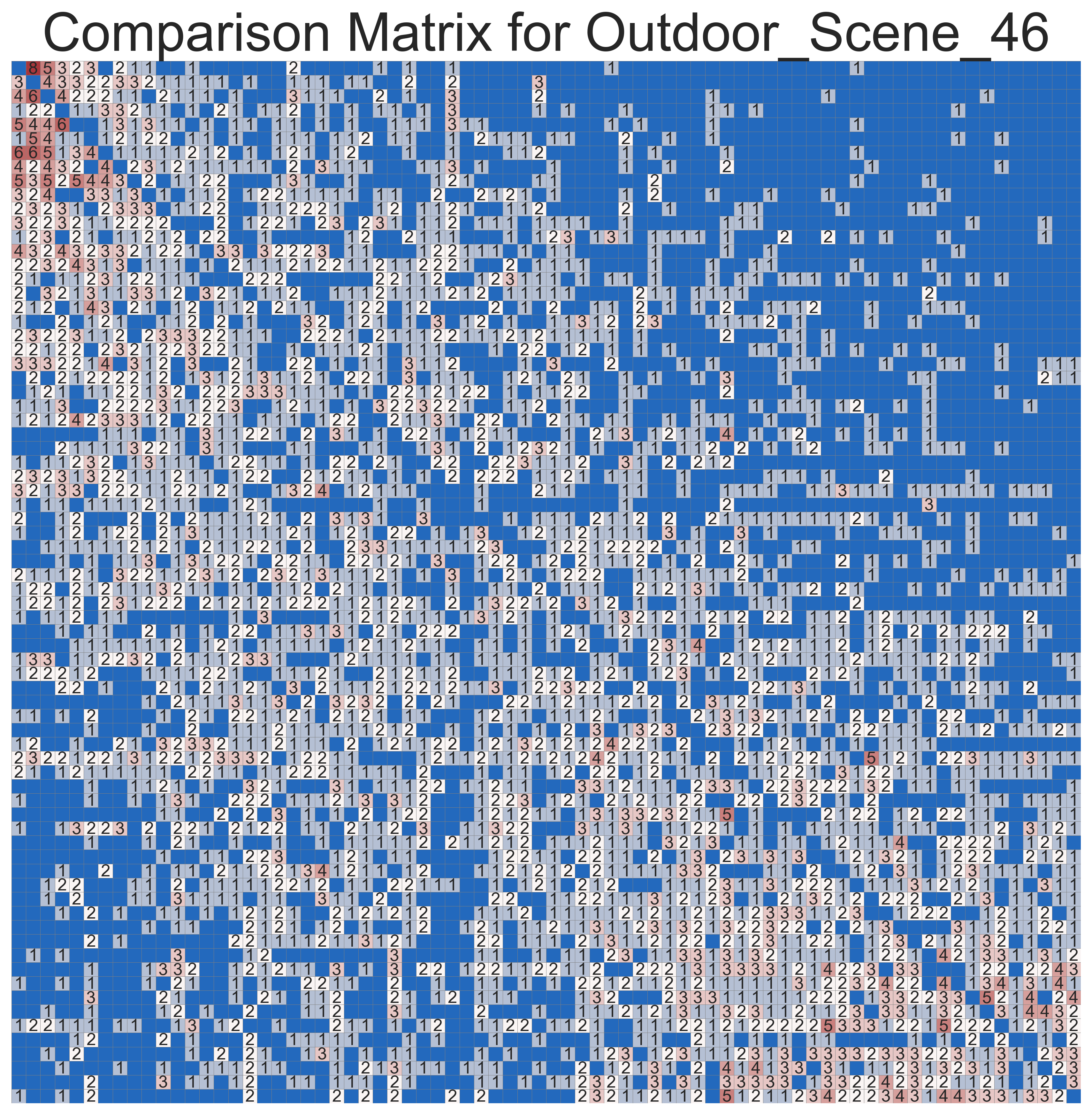} 

    \end{tabular}
 \caption{\footnotesize{Comparison matrices from the PIQ23 dataset, indexed by images with sorted JOD scores. These matrices demonstrate varying levels of sparsity and comparison counts.}}

    \label{fig:COMP_MATS}
\end{figure}

As far as we know, the only available annotated BIQA dataset with pairwise comparisons is PIQ23 \cite{chahine2023image}
The quality scores are generated from sparse comparison matrices using the TrueSkill \cite{herbrich2006trueskill} psychometric scaling algorithm. The public version of PIQ23 does not include the pairwise comparison matrices, but the authors kindly agreed to give us access to these matrices. The matrices will be publicly available on the PIQ23's page \footnote{\url{https://github.com/DXOMARK-Research/PIQ2023}}. The annotations have been conducted by adopting the approach proposed by Mikhailiuk \etal \cite{mikhailiuk2021active}, who propose an efficient active pair selection technique combined with TrueSkill to minimize the experiment cost. Consequently, the collected matrices are typically very sparse (so-called incomplete design) with an emphasis on diagonally neighboring pairs. \Cref{fig:COMP_MATS} illustrates some examples from the dataset, showcasing the sparsity and significant difference in the comparison distributions between different scenes. 
More analysis of the pairwise data is shown in the supplementary material.

\subsection{Generating quality scores from pairwise comparisons}
\label{sub:scaling}
Our pairwise comparison model can either be used in its simplest form, as a comparison tool or to produce precise image quality scales for single or large image sets: 

\begin{itemize}
    \item \textbf{Multi-image inference.} The process of deriving quality scores for multiple images consists of using PICNIQ to generate preference probabilities for a group of image pairs. We then fill the comparison matrix, $M$, by multiplying a predefined average number of comparisons, $c$, by the predicted probability of preference for each pair $p_{ij}$, such as $M_{ij} = c \times p_{ij}, M_{ji} = c \times (1-p_{ij})$. We proceed by applying a psychometric scaling algorithm, TrueSkill in our case, to translate the comparisons into a coherent scale of quality scores. We could of course employ other approaches like Perez \etal \cite{perez2017practical}, who employ a Gaussian distribution as a prior model for image quality.
    \item \textbf{Single image inference.} To assign a quality score to a single image, we propose to compare it against a set of reference images with established scores. This can be achieved by either integrating the new image into an existing comparison matrix and recalculating the scores for all images or by fixing the reference scores and adjusting the new image score to align it with the established scale. 
\end{itemize}

By only predicting the preference likelihood, PICNIQ allows multiple degrees of freedom over the choices of the scaling algorithm, the prior distribution of image quality, and the integration method of single images. In our experiments, we recreate the comparison matrix of each scene using PICNIQ's predictions.
 It's important to note that the choice of image pairs is just as important in inference as it is in training, to ensure consistent scaling. We suggest using active sampling techniques \cite{xu2018hodgerank, mikhailiuk2021active} for optimal pair selection. 

\section{Experiments}

\subsection{Datasets}
We train our model on the PIQ23 dataset \cite{chahine2024generalized}, where we use 35 out of 50 scenes for training and report results on the other 15 scenes, which are uniformly distributed across the different lighting conditions, encompassing around 1486 out of the 5116 images of PIQ23. Previously reported results \cite{chahine2023image, chahine2024generalized} showcase difficulty in the generalization over this dataset, due to the large diversity in image conditions. We demonstrate the capability of our model on the three attributes presented in PIQ23: details, exposure, and overall. Each attribute includes around	100k+ image pairs for training and 50k+ for testing. 

For further experimentation, we submit PICNIQ to the ``Deep Portrait Quality Assessment'' challenge \cite{chahine2024deep}. This challenge tests the generalization power of BIQA models on a private portrait quality dataset composed of 96 single-person scenes of 7 images each, taken with 6 high-quality smartphone images and 1 DSLR capture edited by a professional photographer as the quality reference. This dataset presents quality differences at a close range, which challenges the granularity of the tested BIQA models.

\subsection{Implementation details}
We train PICNIQ on randomly cropped patches of size $1150 \times 1150$. We use Adam stochastic optimization with learning rates between $10^{-7}$ and $10^{-4}$. For smoother training, we adopt different learning rates per module. For instance, we apply a smaller learning rate to the backbone compared to the fully connected layer. Since we have a large set of pairs over a small number of images (i.e. 100k+ pairs over 3k+ images), we apply a strongly decreasing scheduler to smooth the training and avoid early overfitting. 
Since the comparison matrices include an important number of noisy single count comparison pairs (\cref{fig:COMP_MATS}), we have tested multiple thresholds on the minimum number of comparisons between each pair, $\left(1, 2, 3, \text{or } 4\right)$. Our best results have been obtained for thresholds 2 and 3. 
For the model architecture, we adopt DBCNN's deep dual bilinear backbone \cite{zhang2018blind}, pre-trained on KonIQ-10k \cite{hosu2020koniq}. It consists of a frozen CNN pre-trained on synthetic data from the Waterloo \cite{ma2016waterloo} and PASCAL VOC \cite{everingham2010pascal} datasets, combined with a VGG-16 for authentic quality extraction.
Finally, we ensure pair order neutrality by randomly ordering images in input pairs to prevent preference bias (always choosing the first image as best and vice versa). For memory optimization, we have developed an image caching system, which stacks all images present in a single batch and avoids reserving extra memory for images that are repeated in multiple pairs in the same batch. Our experiments have been performed using PyTorch. We have trained our models with either a configuration of 32 Nvidia V100 or 16 Nvidia A100 GPUs, with a batch size of 4 or 10 pairs respectively, with a training time between 10 to 15 hours.

\subsection{Baseline methods}

We compared PICNIQ with several BIQA models trained and tested on PIQ23: DB-CNN \cite{zhang2018blind}, HyperIQA \cite{su2020blindly}, MUSIQ \cite{ke2021musiq}, SEM-HyperIQA \cite{chahine2023image}, FHIQA \cite{chahine2024generalized}, and UNIQUE \cite{zhang2021uncertainty}. DB-CNN, UNIQUE, and two MUSIQ models were pre-trained on the LIVE Challenge, KonIQ-10k, and PaQ-2-PiQ datasets. For all HyperIQA variants, only the backbone was pre-trained on ImageNet without any subsequent IQA-specific pre-training. We note that UNIQUE was re-implemented and trained on the same pairs as PICNIQ, adhering to the architecture and loss recommendations of the original authors.

Due to limited resources, we were unable to perform extensive hyperparameter optimization. Instead, we manually and lightly tuned all nine models directly on the PIQ23 testing set. This approach, while less rigorous than automated optimization methods based on cross-validation, allowed us to achieve reasonable performance within our resource constraints. Additionally, we tested the models on the "Deep Portrait Quality Assessment" challenge \cite{chahine2024deep}, which provides an extra layer of fairness and robustness for benchmarking.

\begin{figure*}[!h] 

    \centering
        \begin{tabular}{ccc}
        \includegraphics[width=0.66\columnwidth]{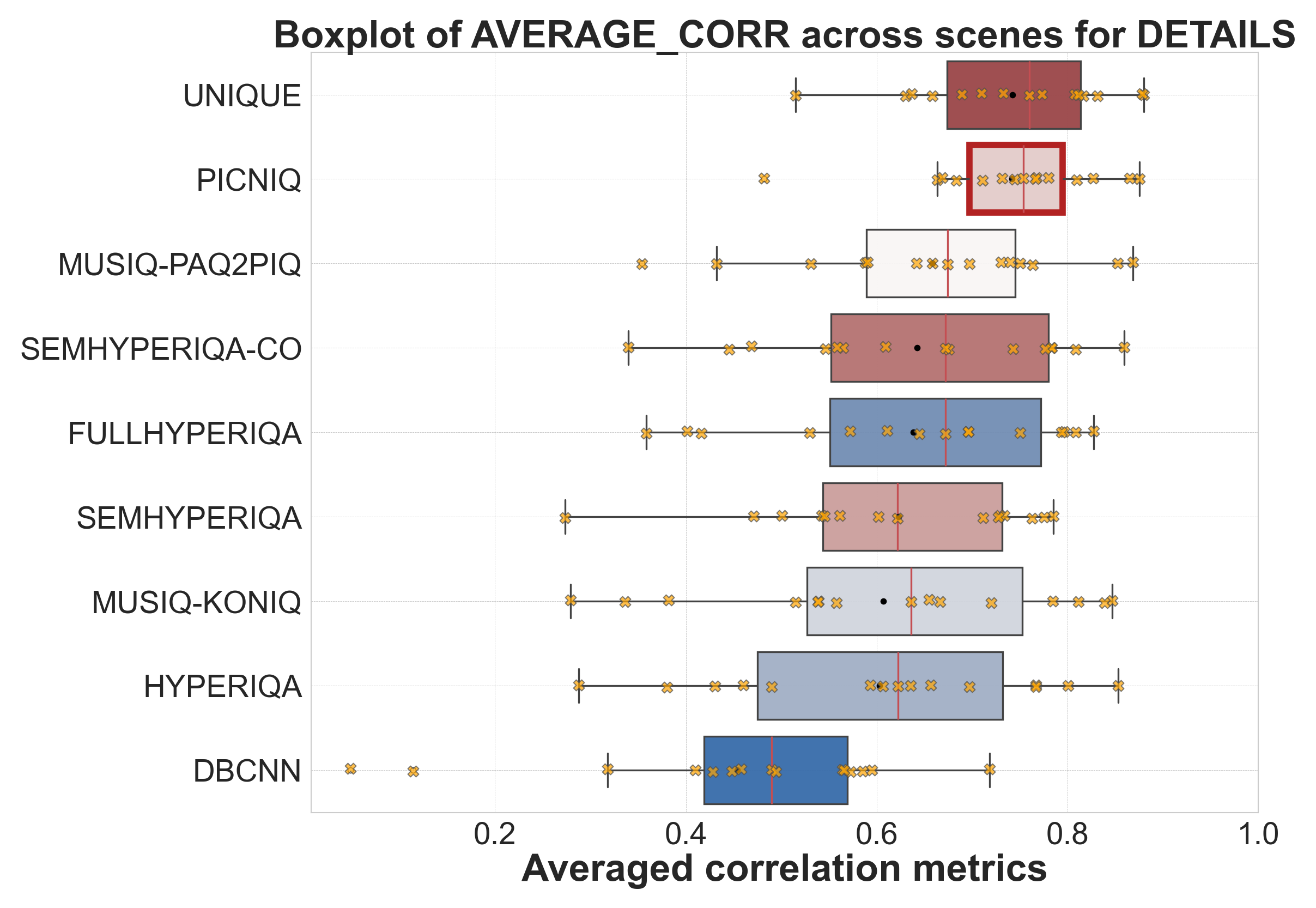} &
        \includegraphics[width=0.66\columnwidth]{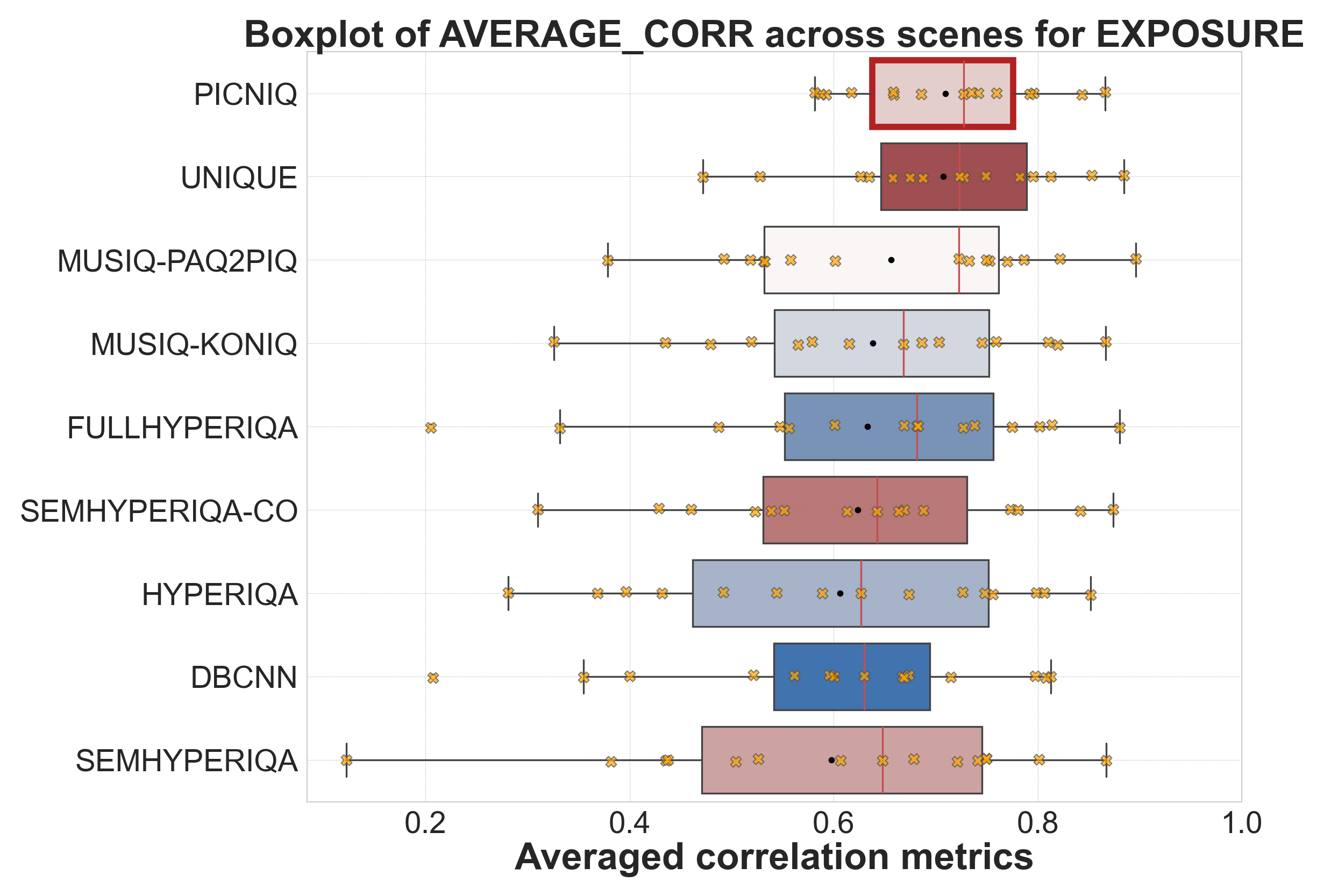} &
        \includegraphics[width=0.66\columnwidth]{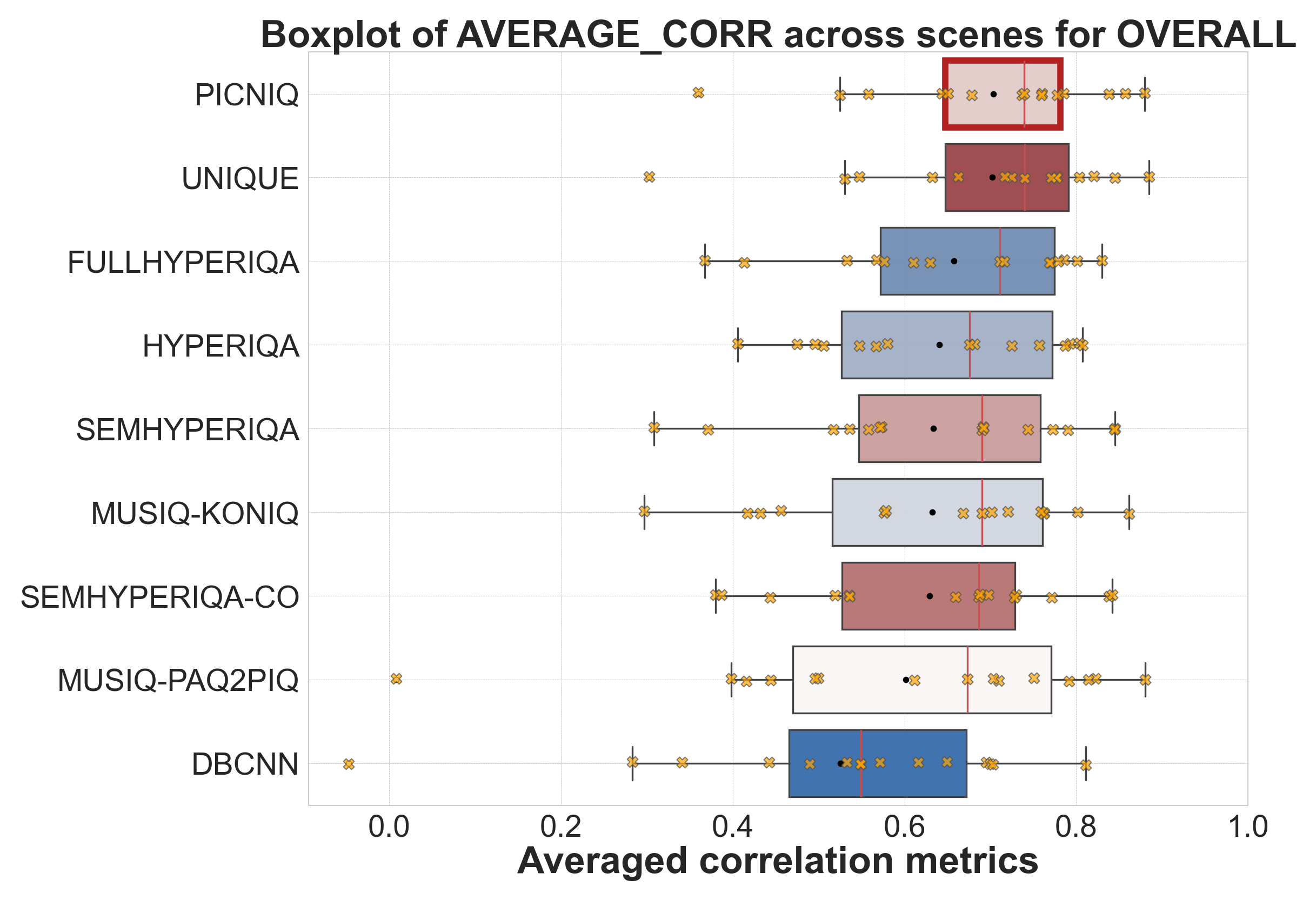}
        \\
        \includegraphics[width=0.66\columnwidth]{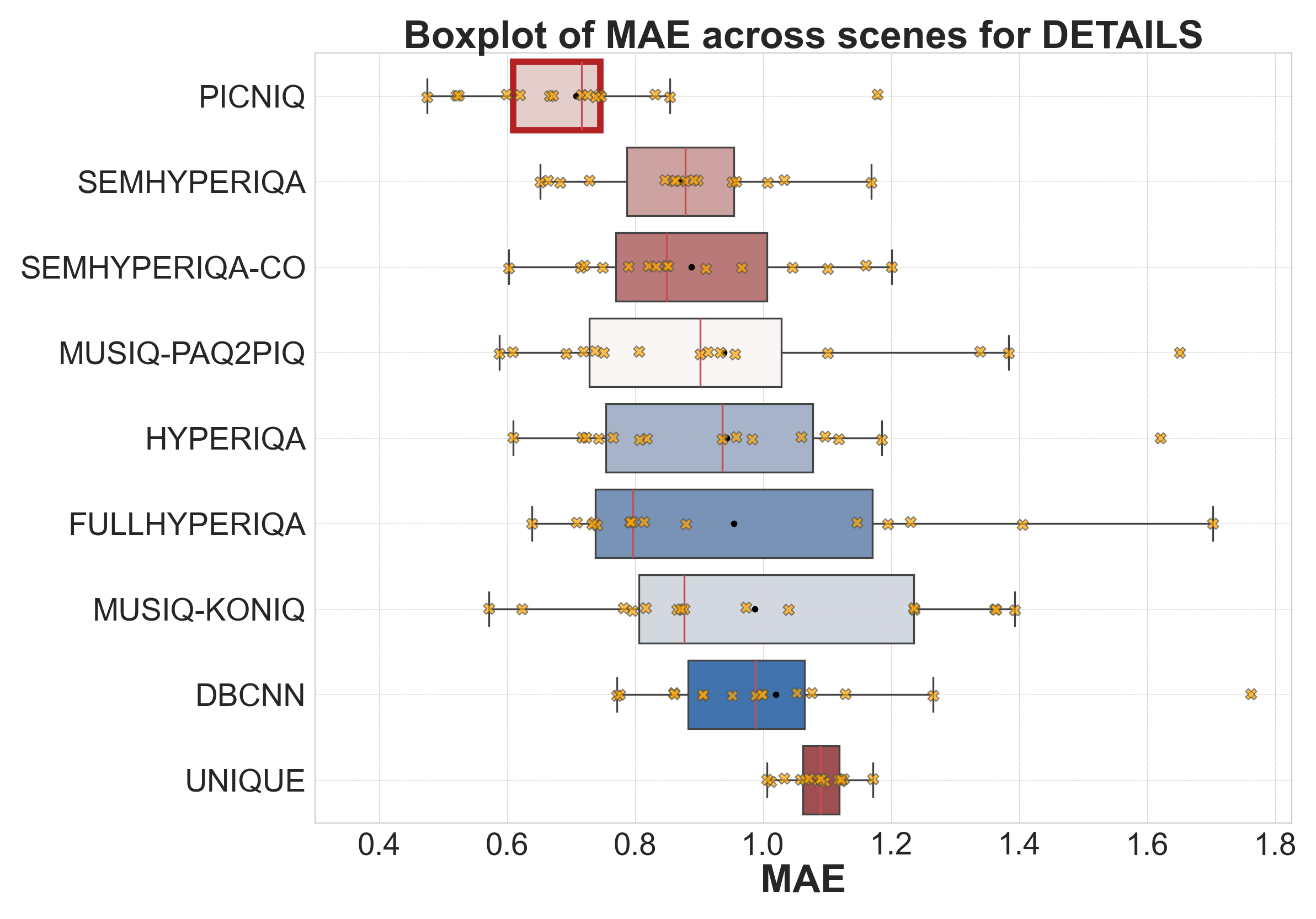} &
        \includegraphics[width=0.66\columnwidth]{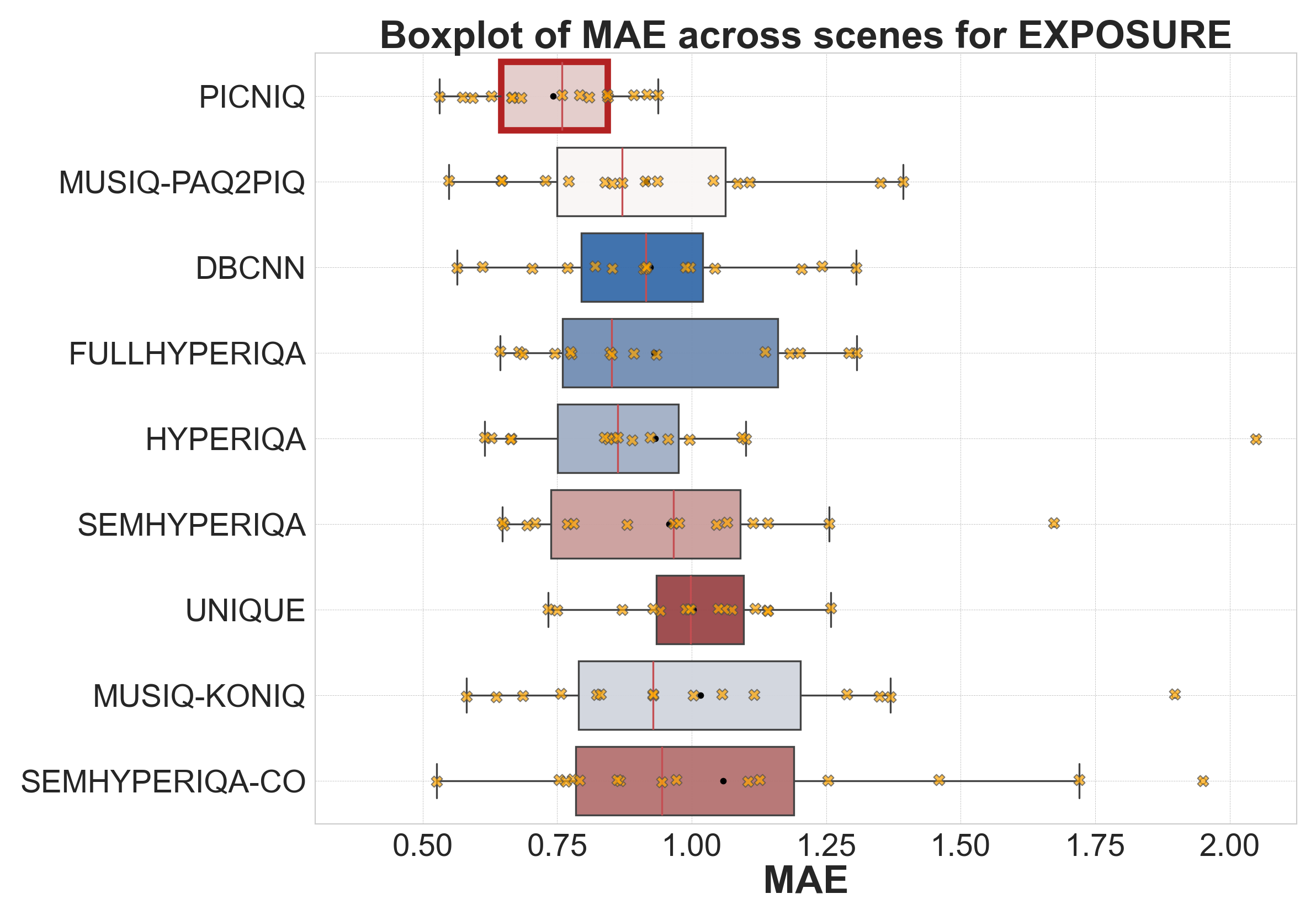} &
        \includegraphics[width=0.66\columnwidth]{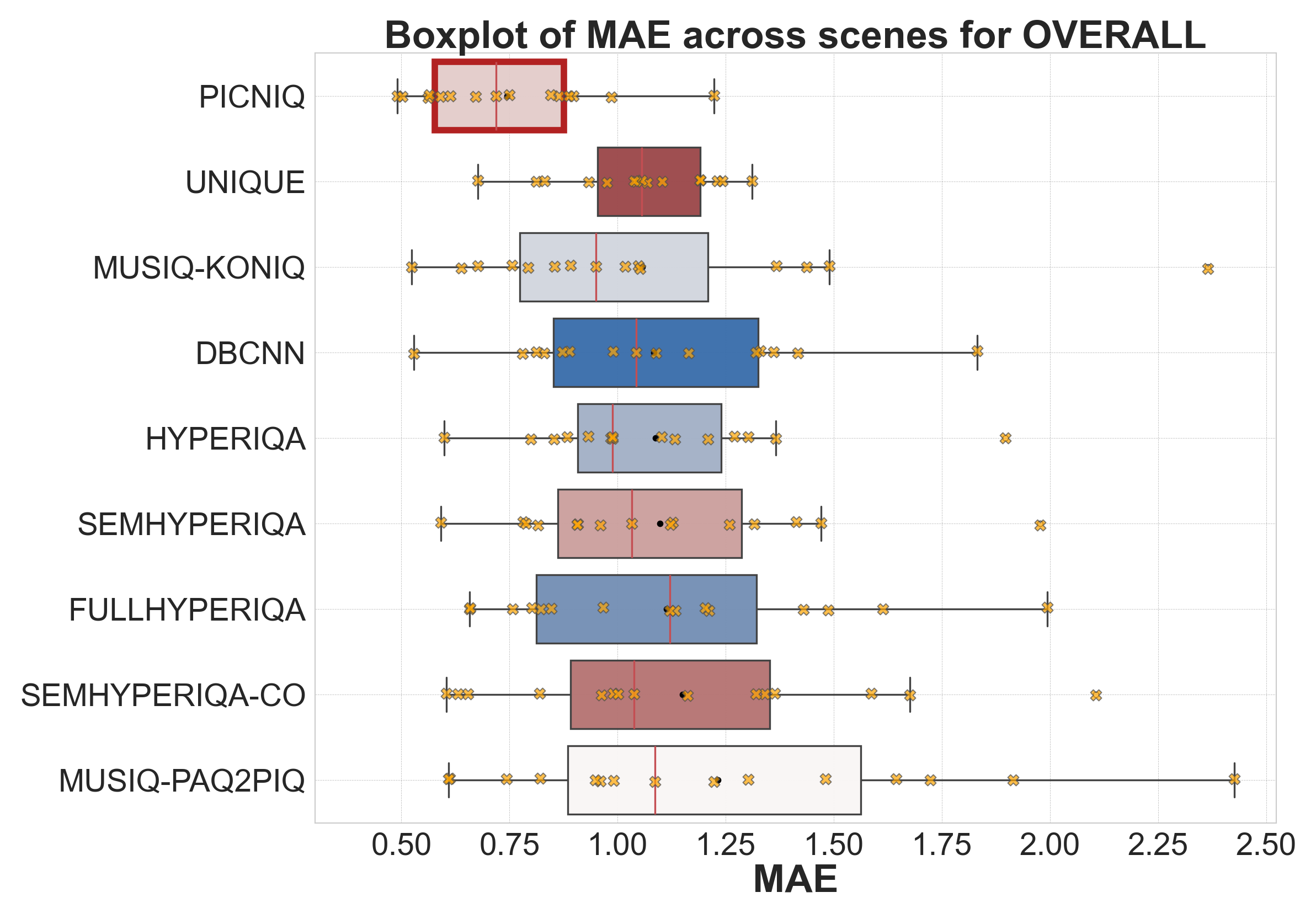} \\
    \end{tabular}
    \caption{\footnotesize{Comparative analysis of IQA models based on the averaged correlation metrics \textit{(top - larger is better)} and mean absolute error \textit{(bottom - smaller is better)} across all scenes and for the three attributes of PIQ23. The results showcase the superiority of PICNIQ over previous models in all metrics.}}
    \label{fig:BOXPLOTS}
\end{figure*}

\begin{figure*}[!h] 
    \setlength{\tabcolsep}{1pt}
    \centering
    \begin{tabular}{ccc}
        {{\includegraphics[width=0.66\columnwidth]{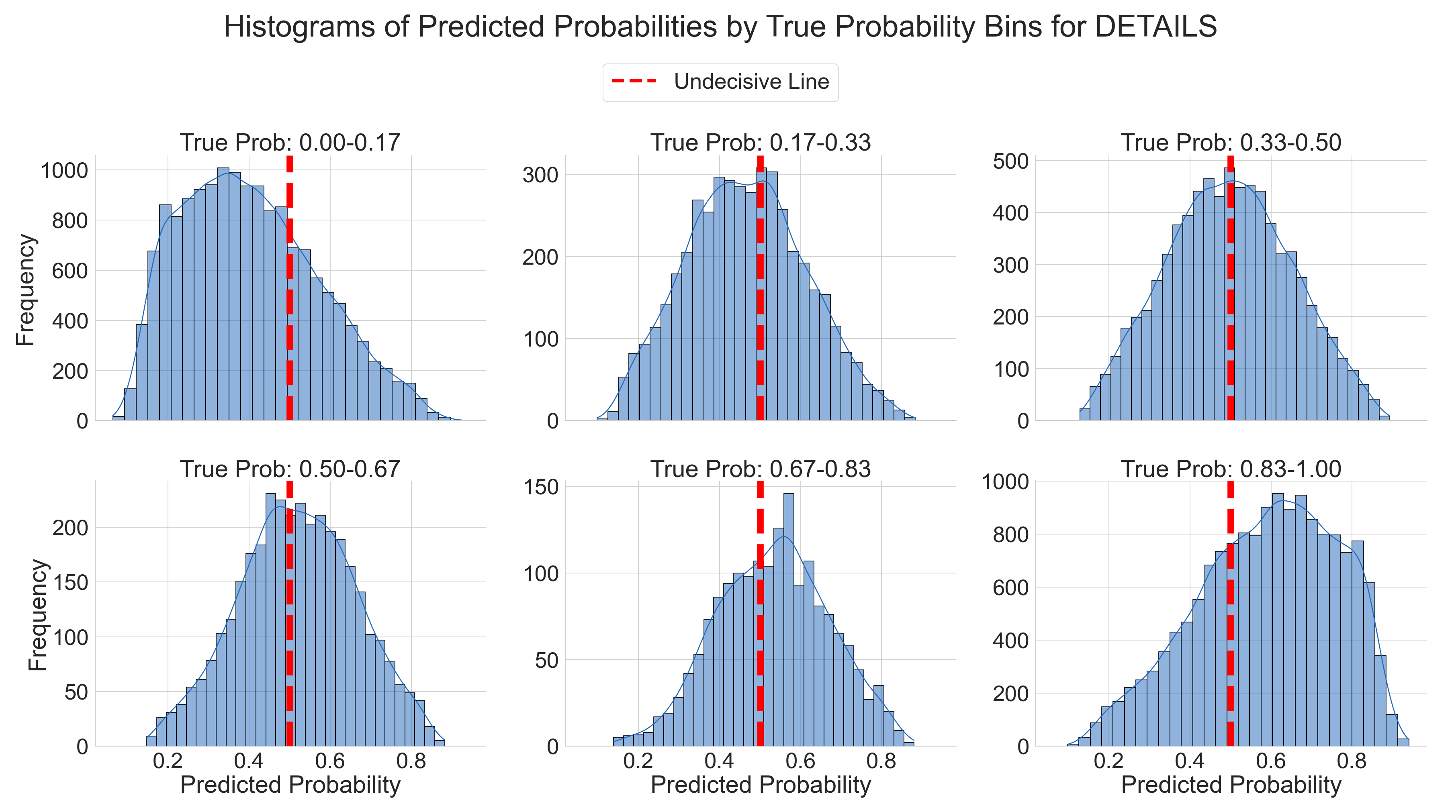} }} &
        {\includegraphics[width=0.66\columnwidth]{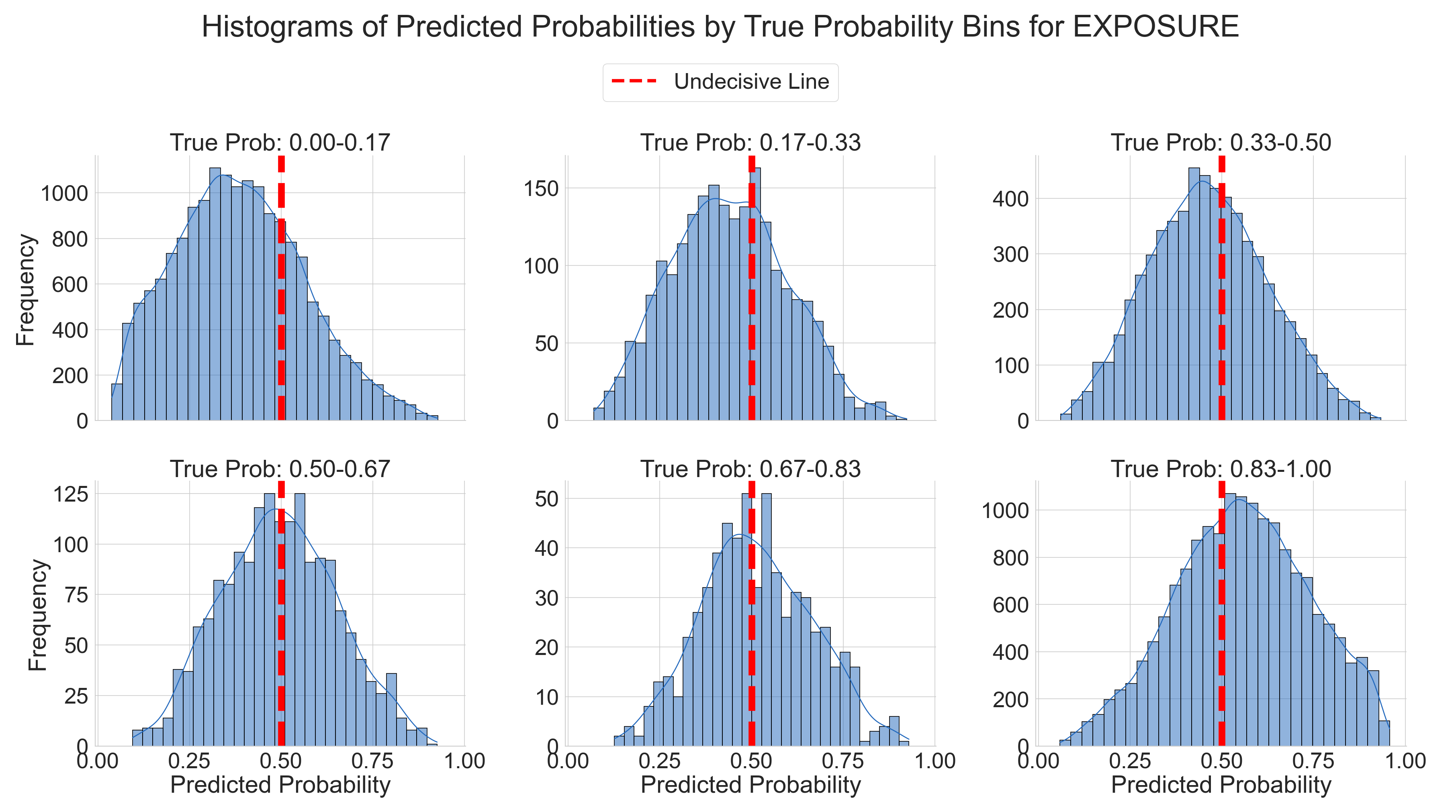}} &
        {\includegraphics[width=0.66\columnwidth]{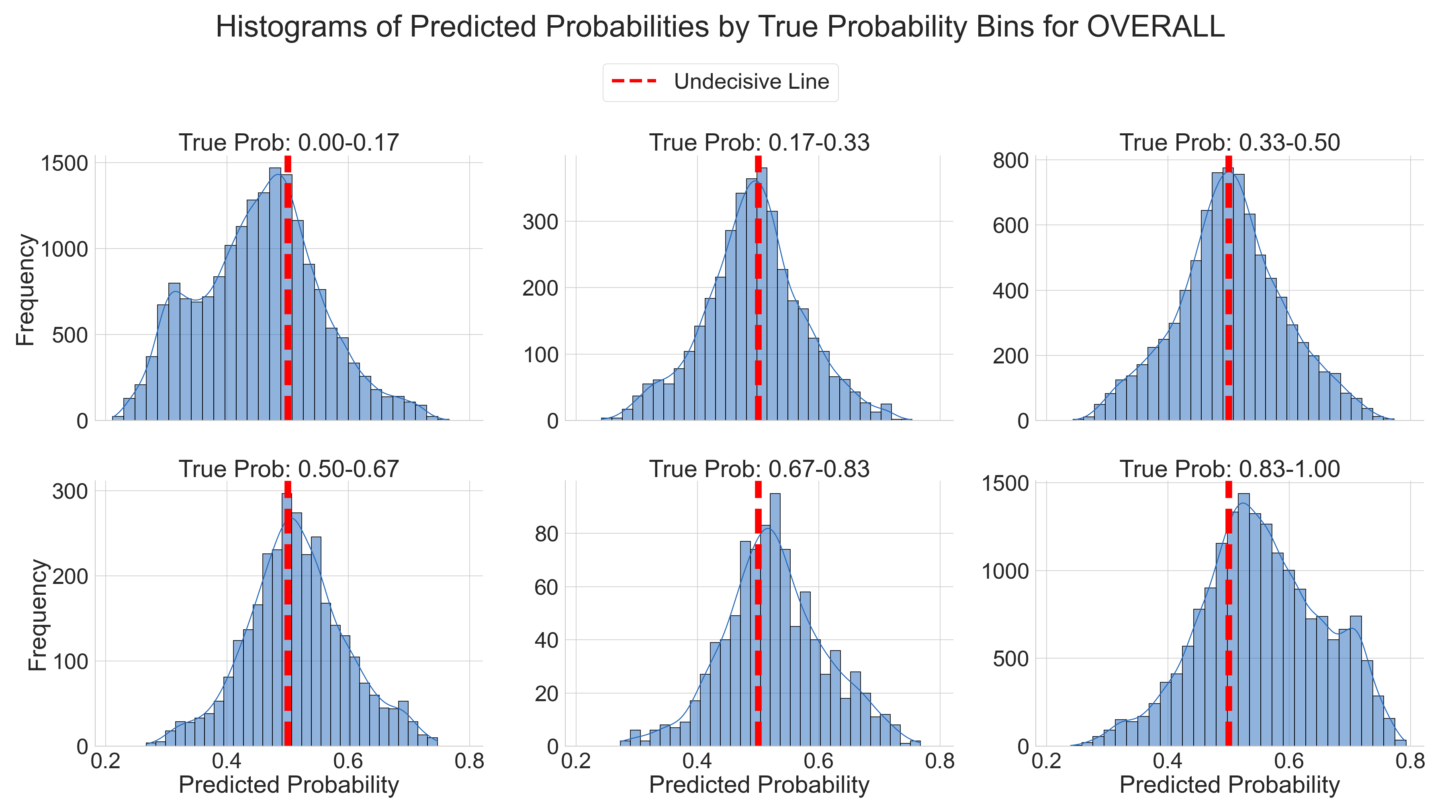}}\\
    \end{tabular}
    \caption{\footnotesize{The histogram of PICNIQ's predictions on PIQ23 separated by their corresponding ground truth values over 6 bins. We can observe that PICNIQ’s predictions are reasonably aligned and calibrated with the ground truth probability, suggesting that the model is increasingly confident in predicting higher probabilities when they are indeed higher. The red dashed line represents the indecisive line with a probability of $0.5$.}}
    \label{fig:HISTS}
\end{figure*}

\subsection{Metrics}
To evaluate performance, we compute Pearson's linear correlation coefficient (PLCC), Spearman’s rank correlation coefficient (SRCC), Kendal's rank correlation (KRCC), the averaged correlations, and the mean absolute error (MAE) between the model outputs and the ground-truth scores. 
In the two testing datasets, each scene is annotated individually. Therefore, we calculate the metrics for each scene separately and then aggregate the performance on each metric across all scenes using the median, $M_{\text{Med}} = M_{(\tfrac{s}{2})}$
where \( s \) denotes the total number of scenes, and \( M_{(i)} \) represents the \( i \)-th smallest scene metric value among the sorted scenes. We also report the mean and the margin of error (MoE) across the scenes. 

\begin{table}[t!] 
\centering
\setlength{\tabcolsep}{5pt}

\resizebox{0.99\columnwidth}{!}{

\begin{tabular}{cl|cccc}
\toprule
\multirow{3}{*}{\#} & \multirow{3}{*}{Model\textbackslash Attribute} & \multicolumn{4}{c}{Details} \\
\cmidrule{3-6}
& & SRCC{\slshape\color{teal}\footnotesize{ (M\textpm MoE)}} & PLCC & KRCC & MAE \\
\midrule
1 & DB-CNN (LIVE C) & 0.59{\slshape\color{teal}\footnotesize{ (0.53\textpm 0.12)}} & 0.51{\slshape\color{teal}\footnotesize{ (0.44\textpm 0.12)}} & 0.45{\slshape\color{teal}\footnotesize{ (0.39\textpm 0.09)}} & {0.99}{\slshape\color{teal}\footnotesize{ (1.02\textpm 0.13)}} \\
2 & MUSIQ (KonIQ-10k) & {0.71}{\slshape\color{teal}\footnotesize{ (0.65\textpm 0.11)}} & 0.67{\slshape\color{teal}\footnotesize{ (0.68\textpm 0.09)}} & {0.52}{\slshape\color{teal}\footnotesize{ (0.49\textpm 0.10)}} & 0.88{\slshape\color{teal}\footnotesize{ (0.99\textpm 0.15)}} \\
3 & MUSIQ (PaQ-2-PiQ) & 0.72{\slshape\color{teal}\footnotesize{ (\underline{0.71\textpm 0.09})}} & {0.77}{\slshape\color{teal}\footnotesize{ (\underline{0.73\textpm 0.07})}} & {0.53}{\slshape\color{teal}\footnotesize{ (\underline{0.54\textpm 0.08})}} & 0.90{\slshape\color{teal}\footnotesize{ (0.94\textpm 0.17)}} \\
4 & HyperIQA* & 0.70{\slshape\color{teal}\footnotesize{ (0.66\textpm 0.10)}} & 0.67{\slshape\color{teal}\footnotesize{ (0.66\textpm 0.09)}} & 0.50{\slshape\color{teal}\footnotesize{ (0.49\textpm 0.09)}} & 0.94{\slshape\color{teal}\footnotesize{ (0.94\textpm 0.14)}} \\
5 & SEM-HyperIQA* & {0.73}{\slshape\color{teal}\footnotesize{ (0.69\textpm 0.10)}} & 0.65{\slshape\color{teal}\footnotesize{ (0.67\textpm 0.06)}} & \underline{0.55}{\slshape\color{teal}\footnotesize{ (0.51\textpm 0.08)}} & {0.88}{\slshape\color{teal}\footnotesize{ (\underline{0.87\textpm 0.08})}} \\
6 & SEM-HyperIQA-CO* & {0.75}{\slshape\color{teal}\footnotesize{ (0.70\textpm 0.09)}} & 0.71{\slshape\color{teal}\footnotesize{ (0.70\textpm 0.08)}} & \underline{0.55}{\slshape\color{teal}\footnotesize{ (0.53\textpm 0.08)}} & {0.85}{\slshape\color{teal}\footnotesize{ (0.89\textpm 0.10)}} \\
7 & FHIQA* & {0.74}{\slshape\color{teal}\footnotesize{ (0.69\textpm 0.09)}} & {0.72}{\slshape\color{teal}\footnotesize{ (0.70\textpm 0.08)}} & \underline{0.55}{\slshape\color{teal}\footnotesize{ (0.52\textpm 0.08)}} & \underline{0.80}{\slshape\color{teal}\footnotesize{ (0.95\textpm 0.17)}} \\
8 & UNIQUE (KonIQ-10k) & \textbf{0.84}{\slshape\color{teal}\footnotesize{ (\textbf{0.80\textpm 0.06)}}} & \underline{0.80}{\slshape\color{teal}\footnotesize{ \textbf{(0.81\textpm 0.05)}}} & \textbf{0.64}{\slshape\color{teal}\footnotesize{ \textbf{(0.62\textpm 0.06)}}} & {1.09}{\slshape\color{teal}\footnotesize{ (1.08\textpm 0.03)}} \\
\rowcolor{lightgray!25}
9 & \textbf{PICNIQ (KonIQ-10k)} & \underline{0.83}{\slshape\color{teal}\footnotesize{ (\textbf{0.80\textpm 0.06})}} & \textbf{0.81}{\slshape\color{teal}\footnotesize{ (\textbf{0.81\textpm 0.05})}} & \textbf{0.64}{\slshape\color{teal}\footnotesize{ (\textbf{0.62\textpm 0.06})}} & \textbf{0.72}{\slshape\color{teal}\footnotesize{ (\textbf{0.71\textpm 0.09})}} \\
\bottomrule
\multicolumn{5}{p{270pt}}{}\\
\end{tabular}
}

\resizebox{0.99\columnwidth}{!}{

\begin{tabular}{cl|cccc}
\toprule
\multirow{3}{*}{\#} & \multirow{3}{*}{Model\textbackslash Attribute} & \multicolumn{4}{c}{Exposure} \\
\cmidrule{3-6}
& & SRCC & PLCC & KRCC & MAE \\
\midrule
1 & DB-CNN (LIVE C) & 0.69{\slshape\color{teal}\footnotesize{ (0.65\textpm 0.11)}} & 0.69{\slshape\color{teal}\footnotesize{ (0.67\textpm 0.10)}} & 0.51{\slshape\color{teal}\footnotesize{ (0.49\textpm 0.09)}} & {0.91}{\slshape\color{teal}\footnotesize{ (\underline{0.92\textpm 0.12})}} \\
2 & MUSIQ (KonIQ-10k) & {0.74}{\slshape\color{teal}\footnotesize{ (0.69\textpm 0.09)}} & {0.70}{\slshape\color{teal}\footnotesize{ (0.71\textpm 0.08)}} & {0.55}{\slshape\color{teal}\footnotesize{ (0.52\textpm 0.08)}} & 0.93{\slshape\color{teal}\footnotesize{ (1.02\textpm 0.19)}} \\
3 & MUSIQ (PaQ-2-PiQ) & \textbf{0.79}{\slshape\color{teal}\footnotesize{ ({0.71\textpm 0.09})}} & {0.78}{\slshape\color{teal}\footnotesize{ ({0.73\textpm 0.08})}} & \underline{0.59}{\slshape\color{teal}\footnotesize{ ({0.54\textpm 0.08})}} & 0.87{\slshape\color{teal}\footnotesize{{(0.92\textpm 0.14})}} \\
4 & HyperIQA* & 0.69{\slshape\color{teal}\footnotesize{ (0.66\textpm 0.10)}} & 0.68{\slshape\color{teal}\footnotesize{ (0.67\textpm 0.11)}} & 0.50{\slshape\color{teal}\footnotesize{ (0.49\textpm 0.09)}} & {0.86}{\slshape\color{teal}\footnotesize{ (0.93\textpm 0.19)}} \\
5 & SEM-HyperIQA* & 0.72{\slshape\color{teal}\footnotesize{ (0.65\textpm 0.12)}} & 0.70{\slshape\color{teal}\footnotesize{ (0.66\textpm 0.11)}} & 0.53{\slshape\color{teal}\footnotesize{ (0.49\textpm 0.10)}} & 0.97{\slshape\color{teal}\footnotesize{ (0.96\textpm 0.15)}} \\
6 & SEM-HyperIQA-CO* & 0.70{\slshape\color{teal}\footnotesize{ (0.67\textpm 0.09)}} & 0.70{\slshape\color{teal}\footnotesize{ (0.69\textpm 0.09)}} & 0.52{\slshape\color{teal}\footnotesize{ (0.51\textpm 0.08)}} & 0.94{\slshape\color{teal}\footnotesize{ (1.06\textpm 0.22)}} \\
7 & {FHIQA}* & {0.76}{\slshape\color{teal}\footnotesize{ (0.69\textpm 0.11)}} & {0.71}{\slshape\color{teal}\footnotesize{ (0.69\textpm 0.10)}} & {0.57}{\slshape\color{teal}\footnotesize{ (0.52\textpm 0.10)}} & \underline{0.85}{\slshape\color{teal}\footnotesize{ (0.93\textpm 0.13)}} \\
8 & UNIQUE (KonIQ-10k) & {0.76}{\slshape\color{teal}\footnotesize{ (\underline{0.76\textpm 0.06})}} & \textbf{0.82}{\slshape\color{teal}\footnotesize{ (\underline{0.77\textpm 0.07})}} & {0.57}{\slshape\color{teal}\footnotesize{ (\underline{0.59\textpm 0.07})}} & 1.00{\slshape\color{teal}\footnotesize{{(1.00\textpm 0.08})}} \\
\rowcolor{lightgray!25}
9 & \textbf{PICNIQ (KonIQ-10k)} & \underline{0.77}{\slshape\color{teal}\footnotesize{ (\textbf{0.77\textpm 0.05})}} & \underline{0.81}{\slshape\color{teal}\footnotesize{ (\textbf {0.79\textpm 0.05})}} & \textbf{0.60}{\slshape\color{teal}\footnotesize{ (\textbf{0.59\textpm 0.05})}} & \textbf{0.76}{\slshape\color{teal}\footnotesize{ (\textbf{0.74\textpm 0.07})}} \\
\bottomrule

\multicolumn{5}{p{270pt}}{}\\
\end{tabular}
}

\resizebox{0.99\columnwidth}{!}{

\begin{tabular}{cl|cccc}
\toprule
\multirow{3}{*}{\#} & \multirow{3}{*}{Model\textbackslash Attribute}  & \multicolumn{4}{c}{Overall} \\
\cmidrule{3-6}
 & & SRCC & PLCC & KRCC & MAE \\
\midrule
1 & DB-CNN (LIVE C) & 0.59{\slshape\color{teal}\footnotesize{ (0.57\textpm 0.13)}} & 0.64{\slshape\color{teal}\footnotesize{ (0.58\textpm 0.12)}} & 0.43{\slshape\color{teal}\footnotesize{ (0.42\textpm 0.10)}} & {1.04}{\slshape\color{teal}\footnotesize{ ({1.08}\textpm 0.18)}} \\
2 & MUSIQ (KonIQ-10k) & {0.76}{\slshape\color{teal}\footnotesize{ ({0.68}\textpm 0.10)}} & {0.75}{\slshape\color{teal}\footnotesize{ ({0.70}\textpm 0.09)}} & {0.57}{\slshape\color{teal}\footnotesize{ ({0.51}\textpm 0.09)}} & \underline{0.95}{\slshape\color{teal}\footnotesize{ (\underline{1.06\textpm 0.26})}} \\
3 & MUSIQ (PaQ-2-PiQ) & 0.74{\slshape\color{teal}\footnotesize{ (0.65\textpm 0.14)}} & 0.74{\slshape\color{teal}\footnotesize{ (0.66\textpm 0.13)}} & 0.54{\slshape\color{teal}\footnotesize{ (0.49\textpm 0.11)}} & 1.09{\slshape\color{teal}\footnotesize{ (1.23\textpm 0.29)}} \\
4 & HYPERIQA* & 0.74{\slshape\color{teal}\footnotesize{ (0.69\textpm 0.08)}} & 0.74{\slshape\color{teal}\footnotesize{ (0.71\textpm 0.07)}} & 0.55{\slshape\color{teal}\footnotesize{ (0.52\textpm 0.08)}} & {0.99}{\slshape\color{teal}\footnotesize{ (1.09\textpm 0.17)}} \\
5 & SEM-HyperIQA* & {0.75}{\slshape\color{teal}\footnotesize{ (0.68\textpm 0.10)}} & {0.75}{\slshape\color{teal}\footnotesize{ (0.70\textpm 0.09)}} & 0.56{\slshape\color{teal}\footnotesize{ (0.52\textpm 0.09)}} & 1.03{\slshape\color{teal}\footnotesize{ (1.10\textpm 0.19)}} \\
6 & SEM-HyperIQA-CO* & {0.74}{\slshape\color{teal}\footnotesize{ (0.68\textpm 0.09)}} & 0.74{\slshape\color{teal}\footnotesize{ (0.70\textpm 0.08)}} & 0.55{\slshape\color{teal}\footnotesize{ (0.51\textpm 0.08)}} & 1.04{\slshape\color{teal}\footnotesize{ (1.15\textpm 0.23)}} \\
7 & {FHIQA*} & {0.78}{\slshape\color{teal}\footnotesize{ ({0.71\textpm 0.09})}} & {0.78}{\slshape\color{teal}\footnotesize{ ({0.73\textpm 0.08})}} & {0.59}{\slshape\color{teal}\footnotesize{ (\underline{{0.54}\textpm 0.08})}} & 1.12{\slshape\color{teal}\footnotesize{ (1.11\textpm 0.21)}} \\
8 & UNIQUE (KonIQ-10k) & \underline{0.80}{\slshape\color{teal}\footnotesize{ (\textbf{0.76\textpm 0.09})}} & \underline{0.81}{\slshape\color{teal}\footnotesize{ (\underline{0.77\textpm 0.08})}} & \underline{0.61}{\slshape\color{teal}\footnotesize{ (\textbf{{0.58}\textpm 0.08})}} & 1.06{\slshape\color{teal}\footnotesize{ \underline{(1.05\textpm 0.10)}}} \\
\rowcolor{lightgray!25}
9 & \textbf{PICNIQ (KonIQ-10k)} & \textbf{0.81}{\slshape\color{teal}\footnotesize{ (\underline{0.75\textpm 0.08})}} & \textbf{0.82}{\slshape\color{teal}\footnotesize{ (\textbf{0.78\textpm 0.07})}} & \textbf{0.62}{\slshape\color{teal}\footnotesize{ (\textbf{0.58\textpm 0.08})}} & \textbf{0.72}{\slshape\color{teal}\footnotesize{ (\textbf{0.74\textpm 0.11})}} \\
\bottomrule

\multicolumn{5}{p{330pt}}{*ImageNet, backbone only; \textbf{Best}; \underline{Second best}; `M' Mean; `MoE' Margin of Error.}\\
\end{tabular}
}

\caption{\footnotesize{Performance metrics of various Blind Image Quality Assessment (BIQA) models on PIQ23. The results are presented as Median (Mean\textpm MoE) for each metric over the test scenes of PIQ23. PICNIQ delivers competitive results in all metrics and for all attributes.
}}

\label{tab:benchmarks}
\end{table}

\begin{table}[!ht]
    \centering
    \begin{subfigure}[b]{\columnwidth}
        \centering
        \setlength{\tabcolsep}{5pt}
        \resizebox{0.99\columnwidth}{!}{
        \begin{tabular}{l|c c c|c c c}
        \toprule
        & \multicolumn{3}{c|}{PIQ23~\cite{chahine2023image}} & \multicolumn{3}{c}{Challenge~\cite{chahine2024deep}} \\
        \cline{2-4} \cline{5-7}
        Method & SRCC & PLCC & KRCC & SRCC & PLCC & KRCC \\
        \toprule
        RQ-Net~\cite{chahine2024deep} & 0.820 & 0.839 & 0.621 & \textbf{0.554} & 0.597 & \underline{0.381} \\
        BDVQA~\cite{chahine2024deep}   & \underline{0.849} & \textbf{0.866} & \underline{0.667} & 0.393 & 0.575 & 0.333 \\
        PQE~\cite{chahine2024deep} & \textbf{0.864} & \underline{0.857} & \textbf{0.690} & 0.411 & 0.544 & 0.333 \\
        MoNet~\cite{chahine2024deep}  & 0.760 & 0.791 & 0.580 & 0.357 & 0.433 & 0.286 \\
        SAR~\cite{chahine2024deep} & 0.828 & 0.855 & 0.651 & 0.304 & 0.453 & 0.238 \\
        HyperIQA & 0.740 & 0.736 & 0.550 & 0.429 & 0.560 & 0.333 \\
        SEM-HyperIQA & 0.749 & 0.752 & 0.558 & 0.518 & \underline{0.605} & 0.333 \\
        FHIQA & 0.778 & 0.784 & 0.586 & {0.536} & \textbf{0.633} & \textbf{0.429} \\
        UNIQUE & 0.803 & 0.806 & 0.611 & 0.482 & \underline{0.631} & 0.333 \\
        \rowcolor{lightgray!25}
        \textbf{PICNIQ} & 0.808 & 0.815 & 0.618 & \underline{0.536} & 0.604 & \textbf{0.429} \\
        \bottomrule
        \end{tabular}
        }
        \caption{Challenge Benchmark. The correlation metrics are calculated per scene, and the final result corresponds to the median of scene-wise metrics.}
    \end{subfigure}
    \par\bigskip
    \begin{subfigure}[b]{\columnwidth}
        \centering
        \setlength{\tabcolsep}{3pt}
        \resizebox{0.99\columnwidth}{!}{
        \begin{tabular}{l c c c c}
        \toprule
        Method & PIQ23~\cite{chahine2023image} & Challenge~\cite{chahine2024deep} & Extra Data & Train Res. \\
        \toprule
        RQ-Net~\cite{chahine2024deep} & 0.751 & \textbf{0.517} & Yes & 224 \\
        PQE~\cite{chahine2024deep}  & \textbf{0.811} & 0.429  & Yes & 384  \\
        BDVQA~\cite{chahine2024deep}  & \underline{0.779} & 0.433 & No & 384  \\
        MoNet~\cite{chahine2024deep}  & 0.710 & 0.368  & No & 384  \\
        SAR~\cite{chahine2024deep}   & 0.777 & 0.315   & No & 224  \\
        HyperIQA & 0.676 & 0.456 & No & 1300 \\
        SEM-HyperIQA &0.690  & 0.501 & No & 1300 \\
        FHIQA & 0.711 & {0.515} & No & 1300 \\
        UNIQUE & 0.740 & 0.492 & No & 1150 \\
        \rowcolor{lightgray!25}
        \textbf{PICNIQ } & 0.747 & \underline{0.516}  & No & 1150 \\
        \bottomrule
        \end{tabular}
        }
        \caption{The final metric for each testing set consists of the median of the scene-wise average of the SRCC, PLCC, and KRCC correlations.}
    \end{subfigure}
    \caption{\footnotesize{Comparison of different methods on the PIQ23 \cite{chahine2023image} and the ``Deep Portrait Quality Assessment'' challenge \cite{chahine2024deep} test datasets. We highlight the \textbf{best} and \underline{second best}. PICNIQ ranks second in the challenge with a close performance to the winner.}}
    \label{tab:NTIRE_benchmark}
\end{table}

\subsection{Results}
\label{sub:results}
\subsubsection{Performance over PIQ23}
To demonstrate the performance of PICNIQ, we draw an extensive quantitative analysis on PIQ23. The results, shown in \Cref{tab:benchmarks}, reveal several key observations. First, PICNIQ delivers superior or competitive results across all metrics, especially in the details and overall attributes. Even for the exposure attribute, PICNIQ delivers consistent results with a higher mean and a smaller Margin of Error (MoE) than competitors. 
Second, although UNIQUE and PICNIQ perform similarly, PICNIQ achieves a distinctly smaller MAE. This indicates that PICNIQ is better suited to reproducing fine-grained quality scales aligned with the ground truth, addressing the domain shift problem more effectively. While regression-based models struggle with the JOD scales of the PIQ23 scenes (reflected by large MAE), PICNIQ's pairwise comparison approach simplifies the task, resulting in precise quality scale reproduction (reflected by a much smaller MAE). This highlights the advantage of a ranking-based BIQA metric with score generation as a post-processing step. 

The averaged correlations and MAE distribution across the 15 test scenes of PIQ23 (\Cref{fig:BOXPLOTS}) further support our claims. The boxplots reveal some outlier scenes where metrics are worse than the average, which justifies our choice of using the median to measure performance instead of the mean. PICNIQ outperforms other models with stable performance across all test scenes. Notably, despite MUSIQ appearing to perform well on paper over the exposure attribute, a deeper analysis shows that PICNIQ is significantly more consistent. Furthermore, compared to UNIQUE, which delivers a competitive performance, PICNIQ demonstrates greater stability across different scenes.

\subsubsection{Performance over challenge data}
We have also submitted PICNIQ and UNIQUE to the ``Deep Portrait Quality Assessment'' challenge \cite{chahine2024deep} for additional unbiased testing. As shown in \Cref{tab:NTIRE_benchmark}, PICNIQ ranks slightly below the winning solution, demonstrating strong performance in unseen conditions and good generalization capabilities. It is important to note that competing methods in the challenge used boosting, ensembling techniques, complex architectures, and additional pre-training. Despite its simpler training and architecture strategies, PICNIQ generalizes as well as some of the complex methods and consistently outperforms others. 

\subsubsection{Calibration of the preference predictions}
The histograms of PICNIQ's predictions on PIQ23 (\Cref{fig:HISTS}) show that the model's predictions align well with the ground truth probabilities. The peaks of the histograms generally align with the center of the true probability bins, indicating that the model is fairly well-calibrated. As the true probability increases, the peak of the histogram shifts to the right, showing increased confidence in predicting higher probabilities. However, the model struggles in gray areas, where comparisons are close but a preference choice can still be made. This is likely due to an imbalance in the training dataset towards decisive pairs, with more 0s and 1s than middle values. For a more comprehensive understanding of the training biases, we have included an analysis of the distribution of ground truth probabilities of PIQ23 in the supplementary material.

Overall, the results of our model are consistently better than the previous state-of-the-art, indicating a solid foundation for inspiring future solutions in BIQA for domain shift and domain generalization.

\section{Conclusion}
This paper introduces PICNIQ, a novel pairwise comparison framework, adapted to sparse comparison settings and designed to counter the problem of domain shift and uncertainty in BIQA. Instead of direct quality prediction, PICNIQ is trained to predict the preference likelihood between a pair of images, which is then combined with psychometric scaling algorithms to generate quality scores. PICNIQ can be used as a quality comparison tool or to generate quality scores for a large set of images. We also explore and exploit the pairwise comparison matrices in the PIQ23 dataset, which are made public for future research in BIQA. Through extensive experiments on the generalization split of PIQ23, we demonstrate the high performance of PICNIQ and the granular quality scales it reproduces. Finally, we note that while PICNIQ employs a simple VGG-16 backbone, it outperforms other complex architectures, which opens a wide door for future adaptations that exploit deep architectures and self-supervised models. We also believe that our comparison framework has the potential to become a standard for high-precision digital camera quality assessment metrics, encouraging more pairwise comparison adoption in the IQA literature.

\section{Acknowledgments}
This work was funded in part by the French government under the management of Agence Nationale de la Recherche as part of the “Investissements d’avenir” program, reference ANR-19-P3IA-0001 (PRAIRIE 3IA Institute), the Louis Vuitton/ENS chair in artificial intelligence and the Inria/NYU collaboration.  This work was performed using HPC resources from GENCI-IDRIS (Grant 2023-AD011013850). NC was supported in part by a DXOMARK/PRAIRIE CIFRE Fellowship. Certain sections of this document were improved with the assistance of AI, specifically GPT-4 and Grammarly AI.

\clearpage

{\small
\bibliographystyle{ieee_fullname}
\bibliography{egbib}
}

\end{document}


\title{Appendix for "Pairwise Comparisons Are All You Need"}
\author{}

\maketitle

\section*{Appendix}
This appendix presents multiple examples of the preference predictions of PICNIQ over PIQ23. We display the comparison matrices of some scenes and attributes in PIQ23. Finally, we plot the probability distribution for the different scenes and attributes of the PIQ23 scene split.

\begin{figure}[!htbp]
    \centering
    \setlength{\tabcolsep}{2pt}
    \begin{tabular}{cc}
        \includegraphics[width=0.49\columnwidth]{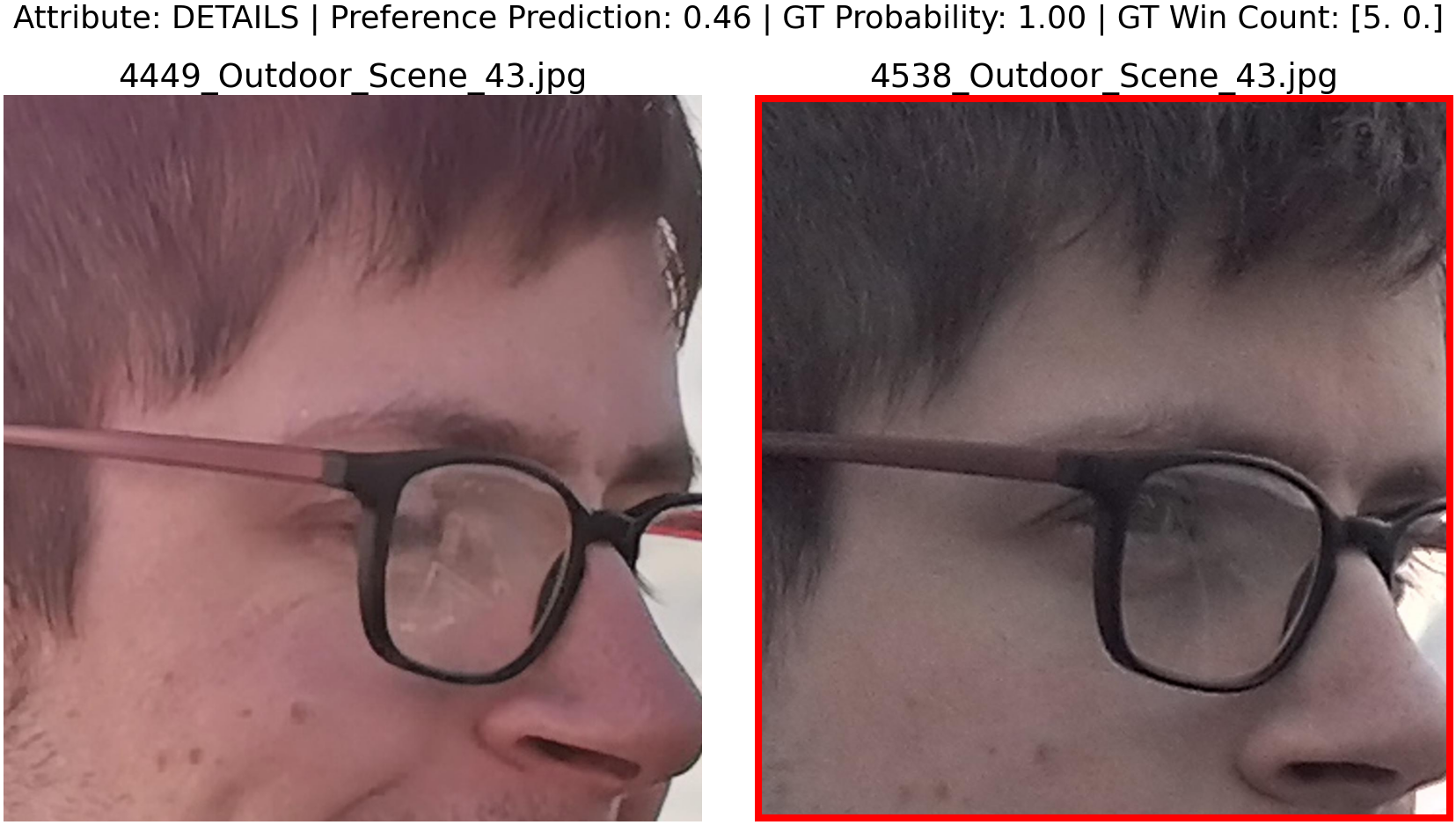} &
        \includegraphics[width=0.49\columnwidth]{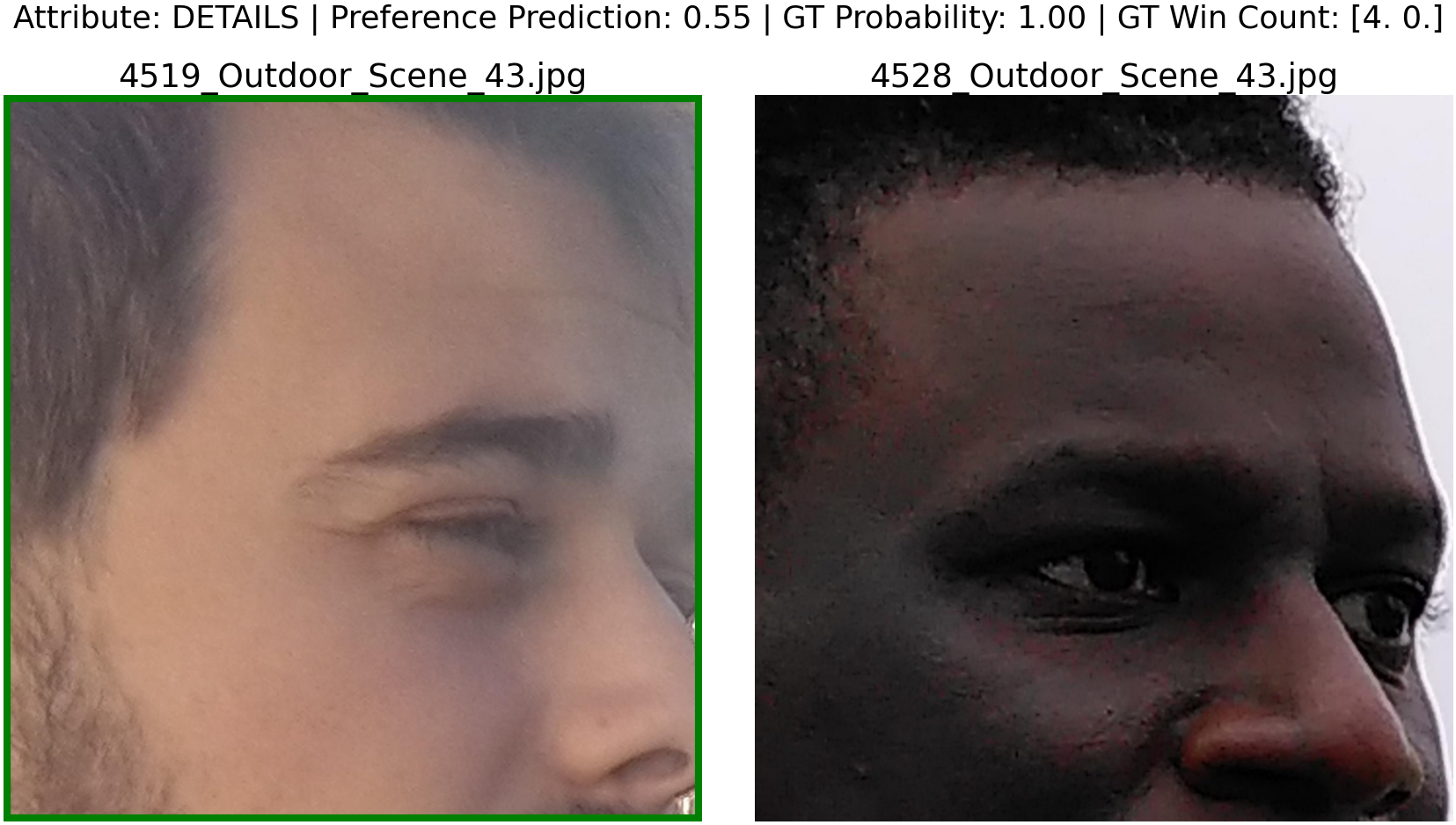} \\
        \includegraphics[width=0.49\columnwidth]{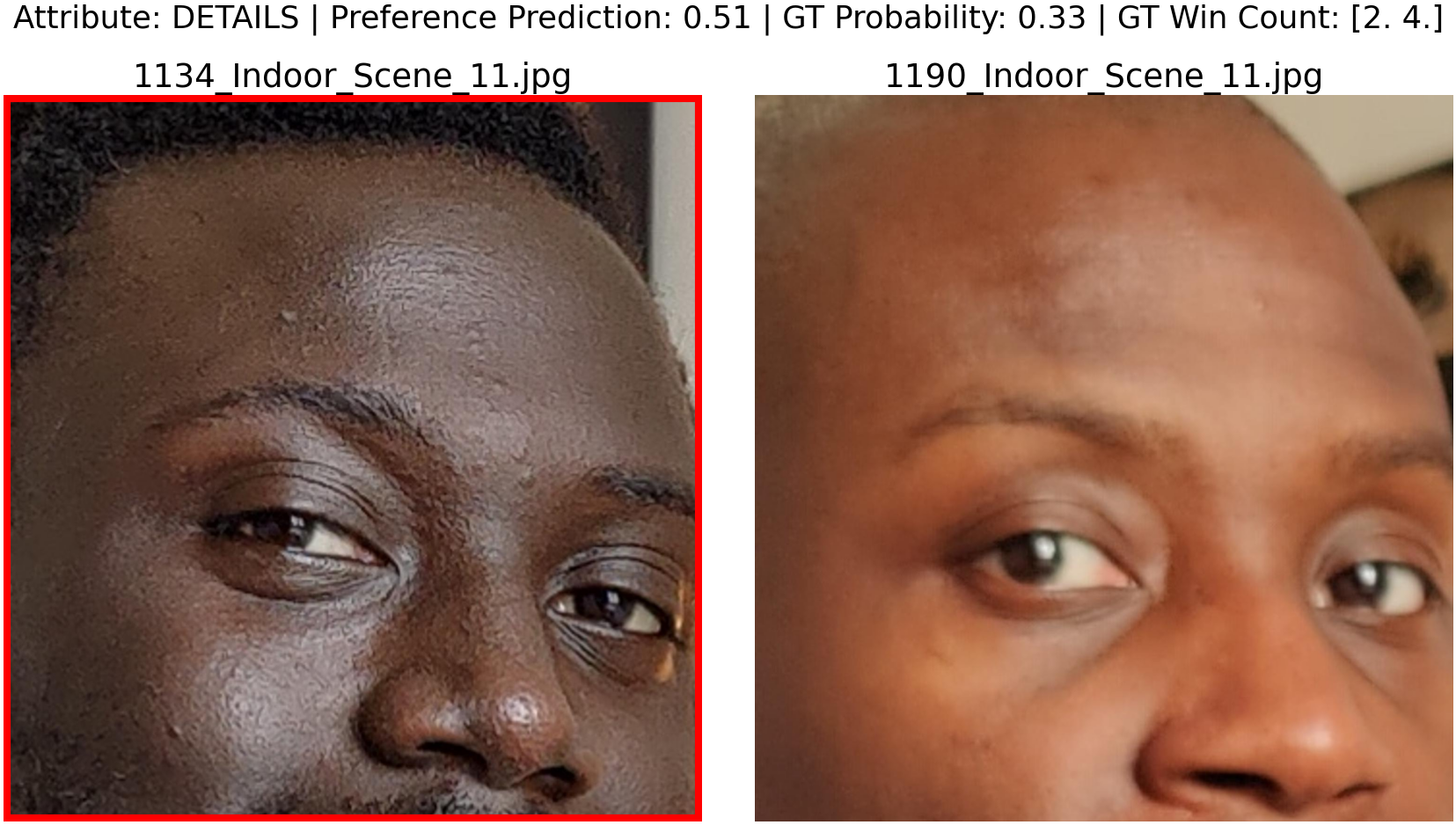} &
        \includegraphics[width=0.49\columnwidth]{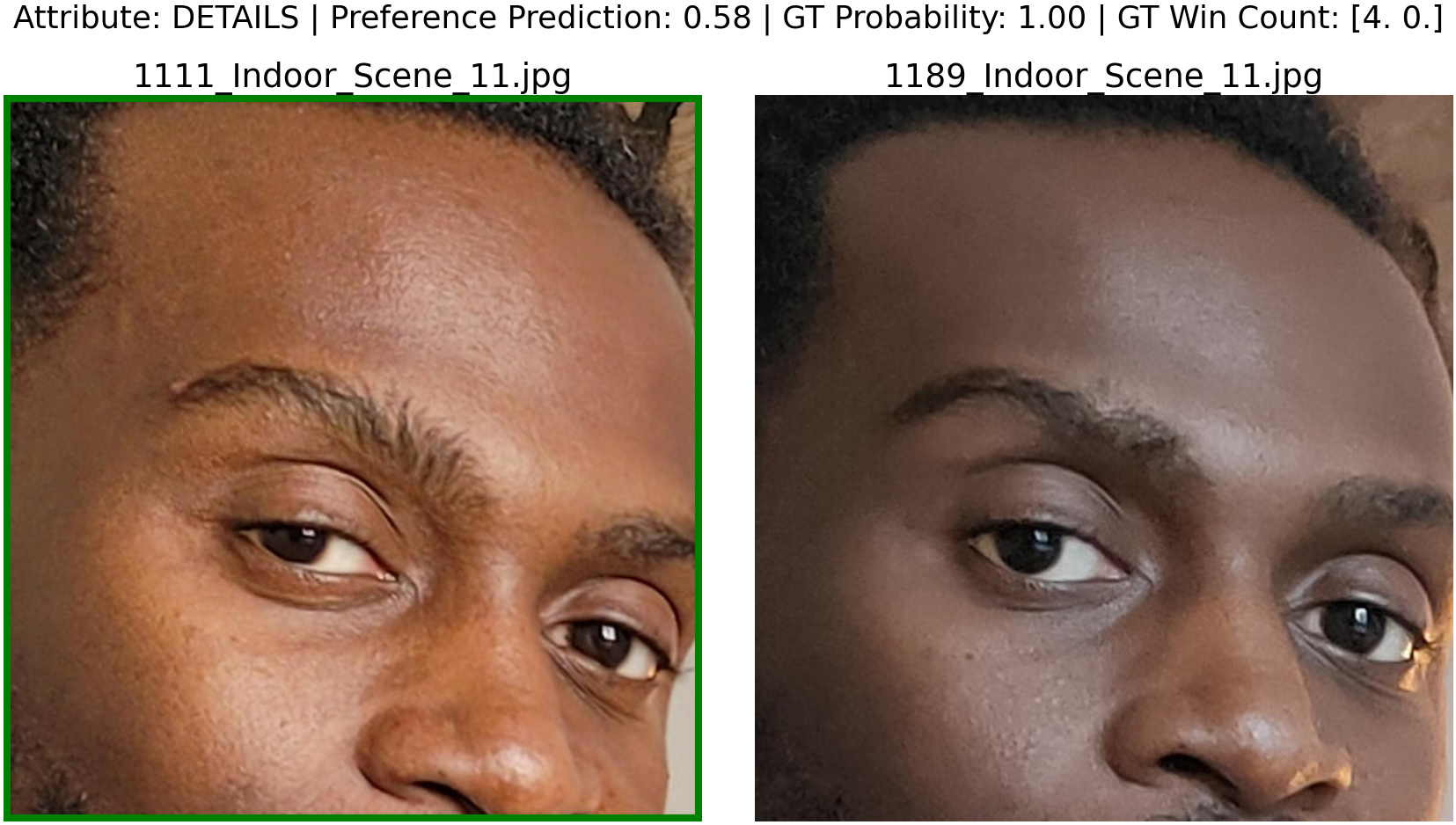} \\
        \includegraphics[width=0.49\columnwidth]{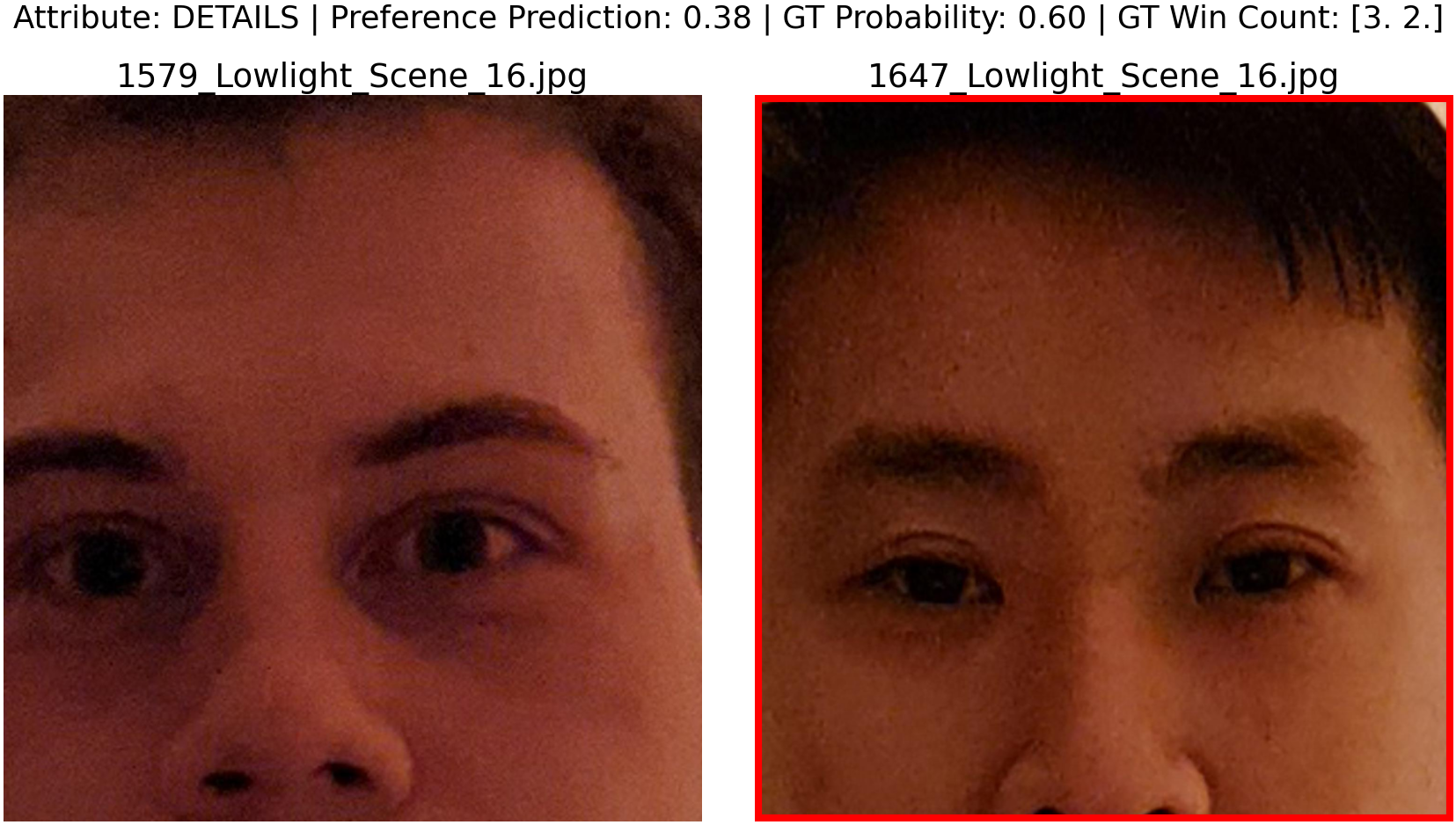} &
        \includegraphics[width=0.49\columnwidth]{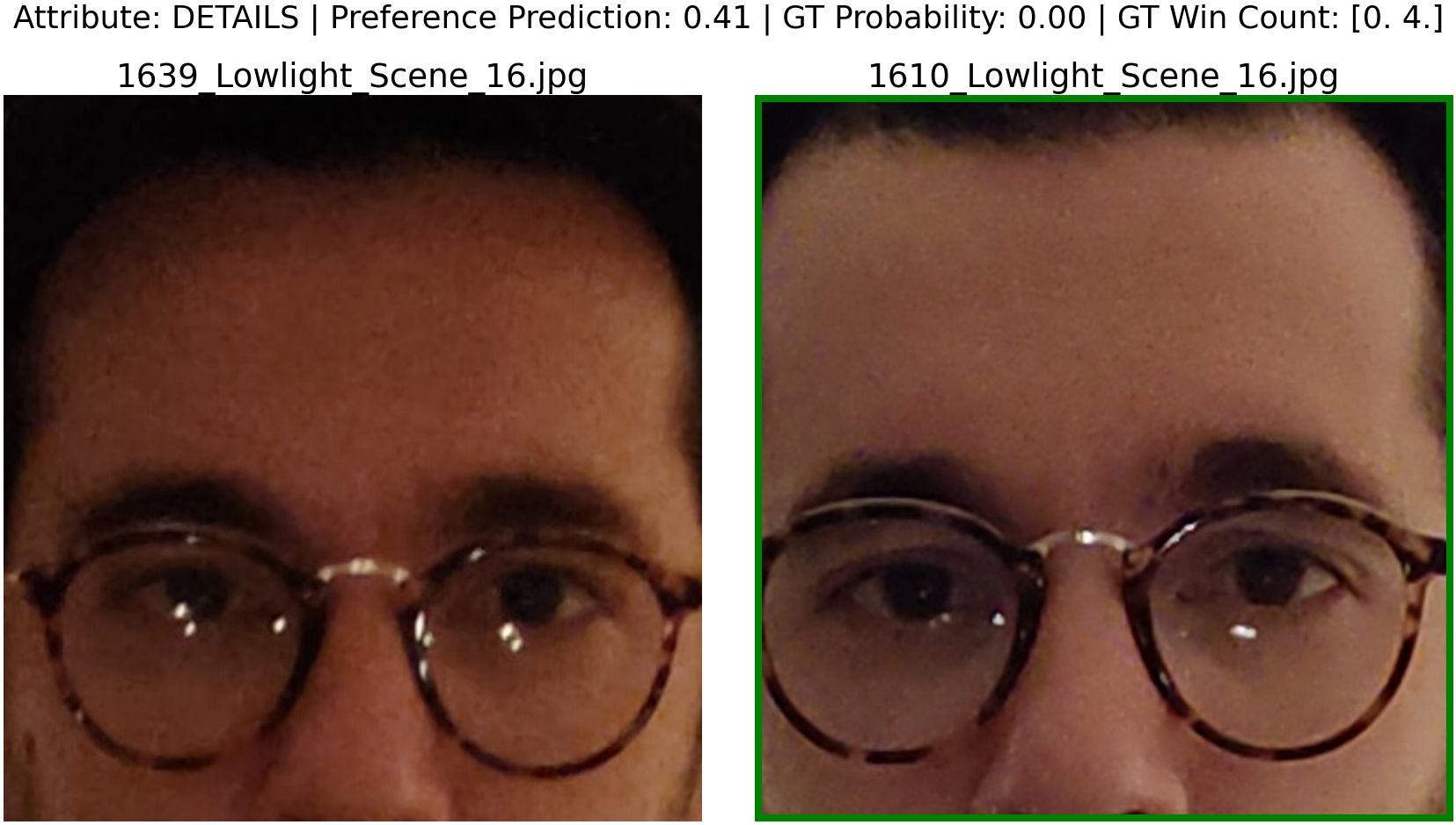} \\
        \includegraphics[width=0.49\columnwidth]{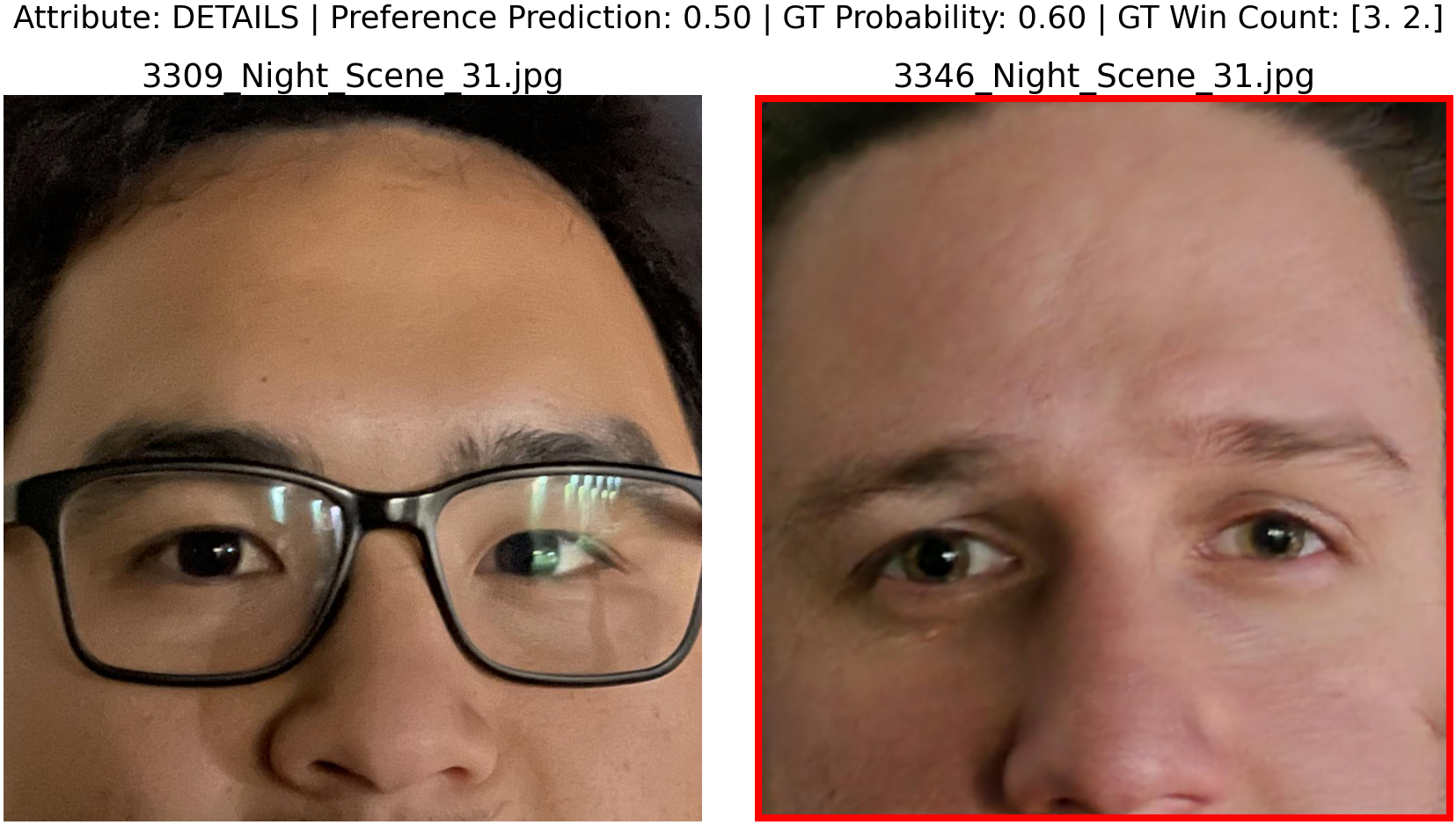} &
        \includegraphics[width=0.49\columnwidth]{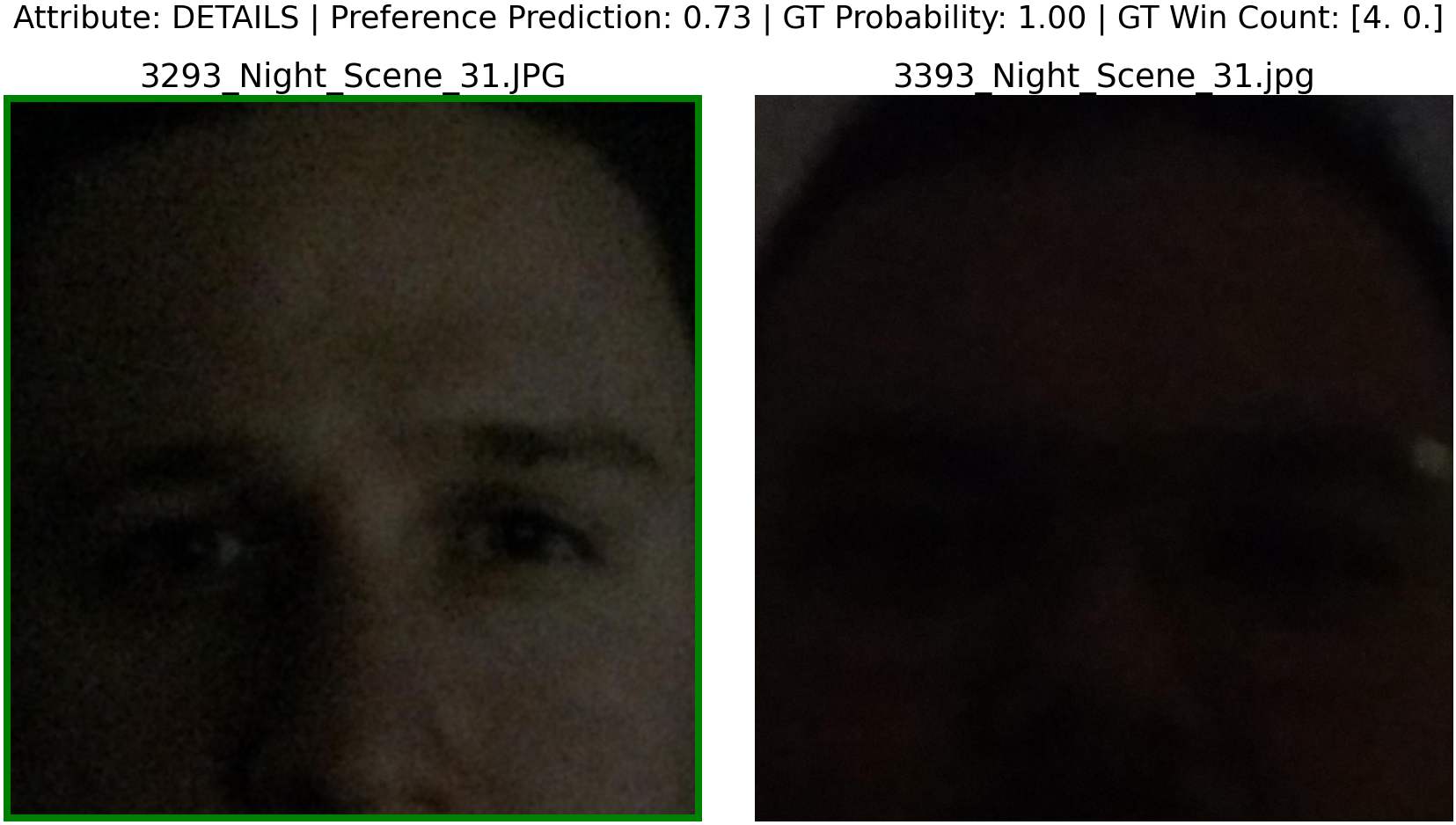} \\
        
    \end{tabular}
    \caption{\footnotesize{Examples of incorrect (left) and correct (right) predictions on multiple scenes for the Details attribute of the PIQ23 dataset. PICNIQ consistently delivers logical and precise comparisons despite the difficulty of the example. Some incorrect predictions are subject to interpretation.}}
    \label{fig:comp_mats_scenes_attributes}
\end{figure}

\begin{figure}[!htbp]
    \centering
    \setlength{\tabcolsep}{2pt}
    \begin{tabular}{cc}
        \includegraphics[width=0.49\columnwidth]{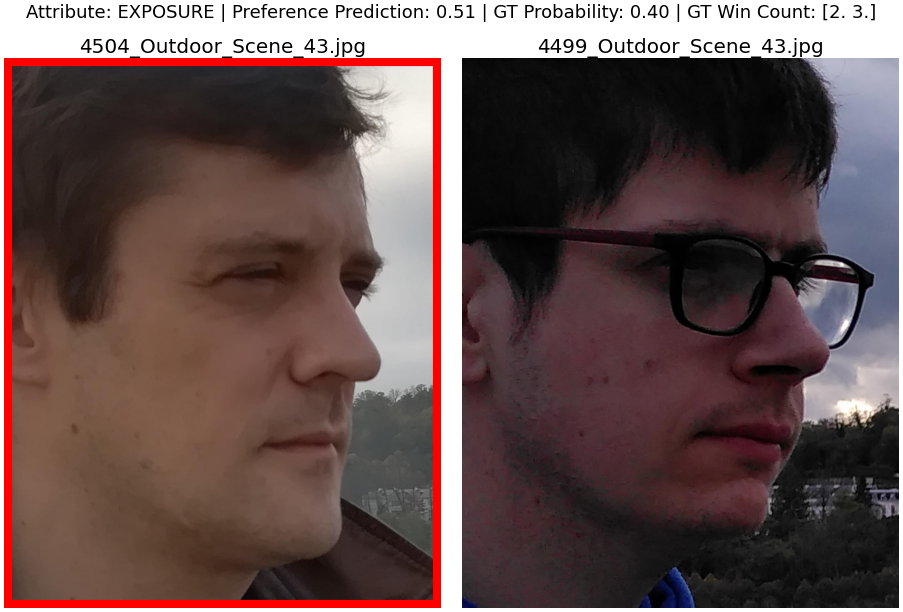} &
        \includegraphics[width=0.49\columnwidth]{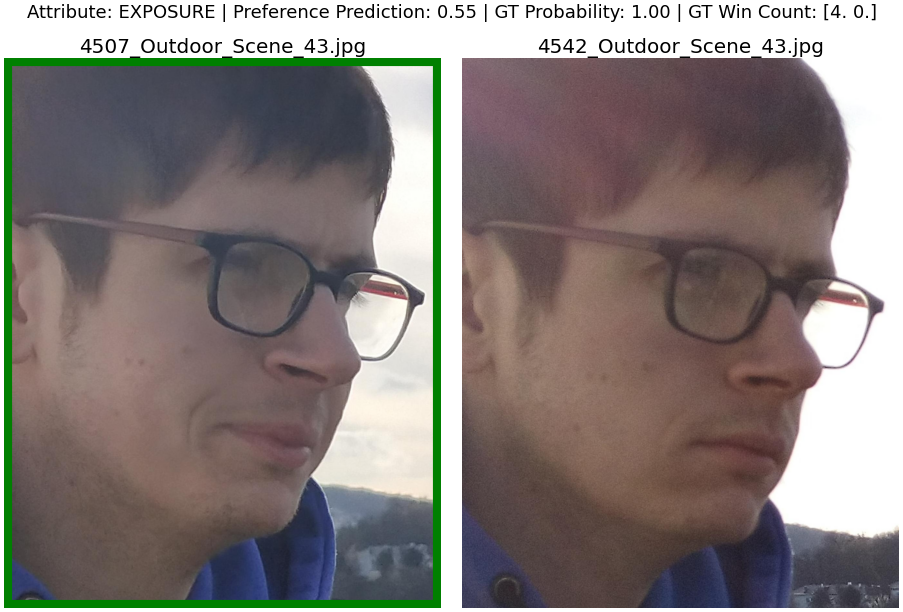} \\
        \includegraphics[width=0.49\columnwidth]{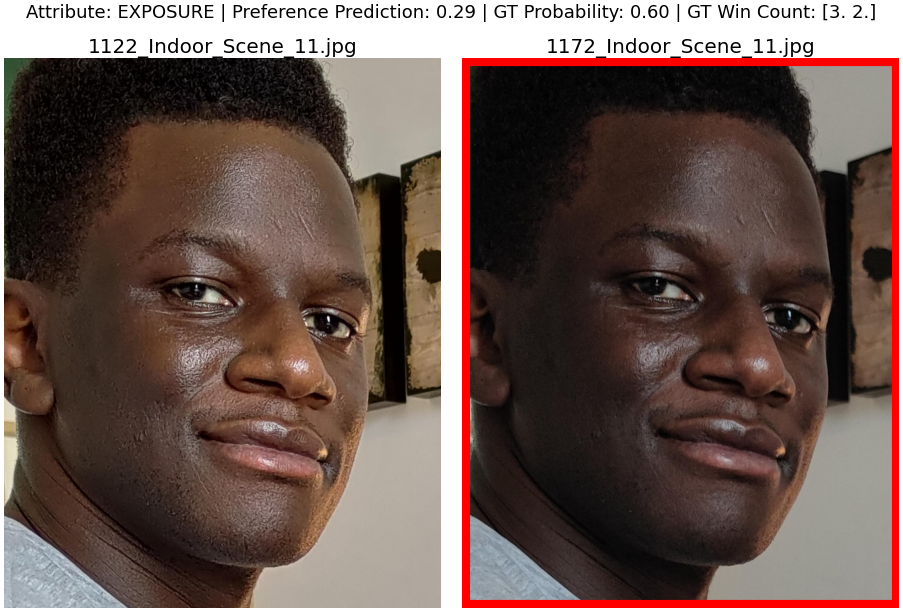} &
        \includegraphics[width=0.49\columnwidth]{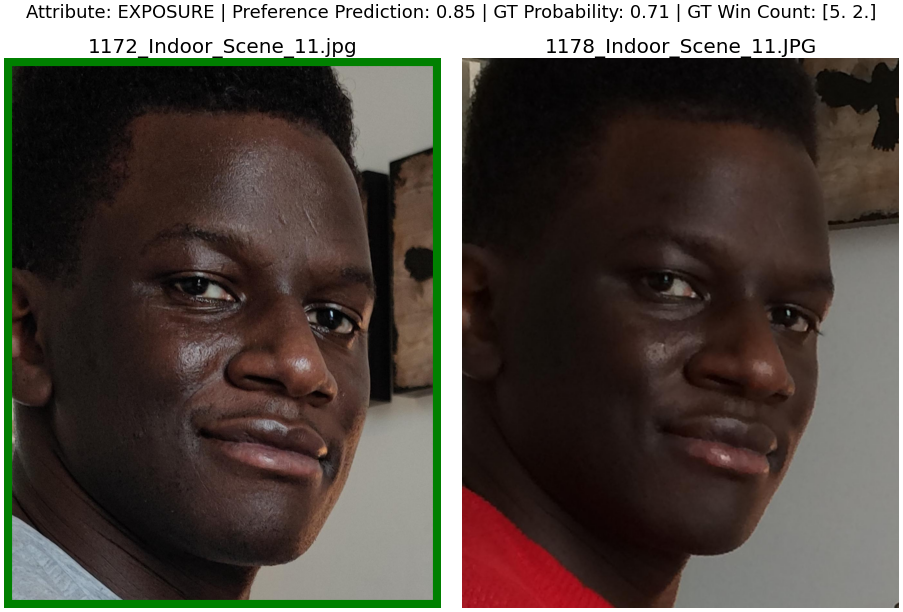} \\
        \includegraphics[width=0.49\columnwidth]{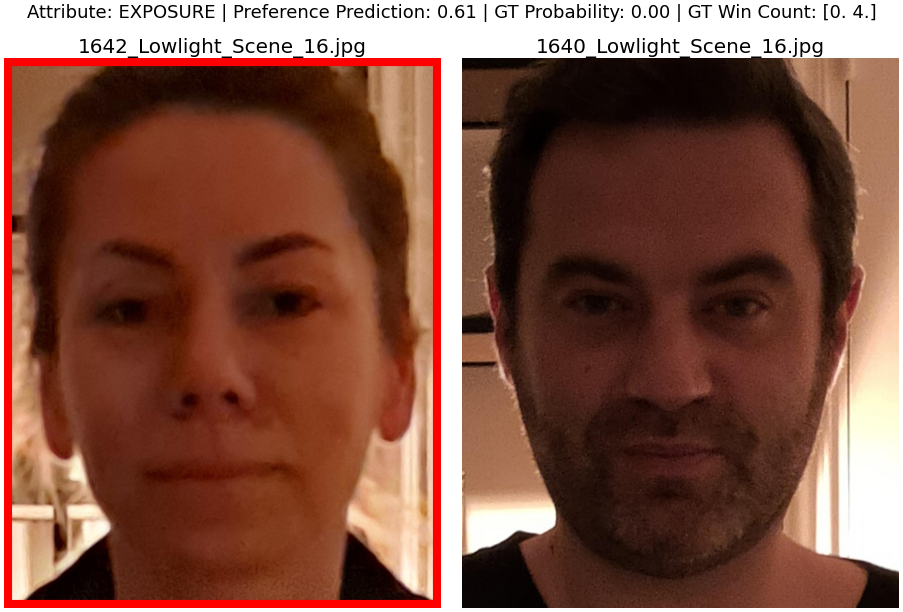} &
        \includegraphics[width=0.49\columnwidth]{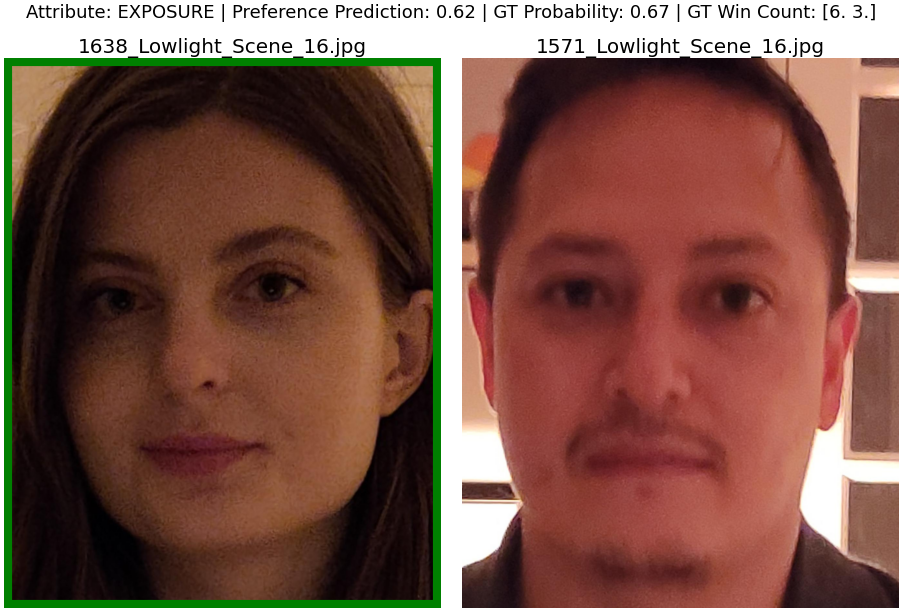} \\
        \includegraphics[width=0.49\columnwidth]{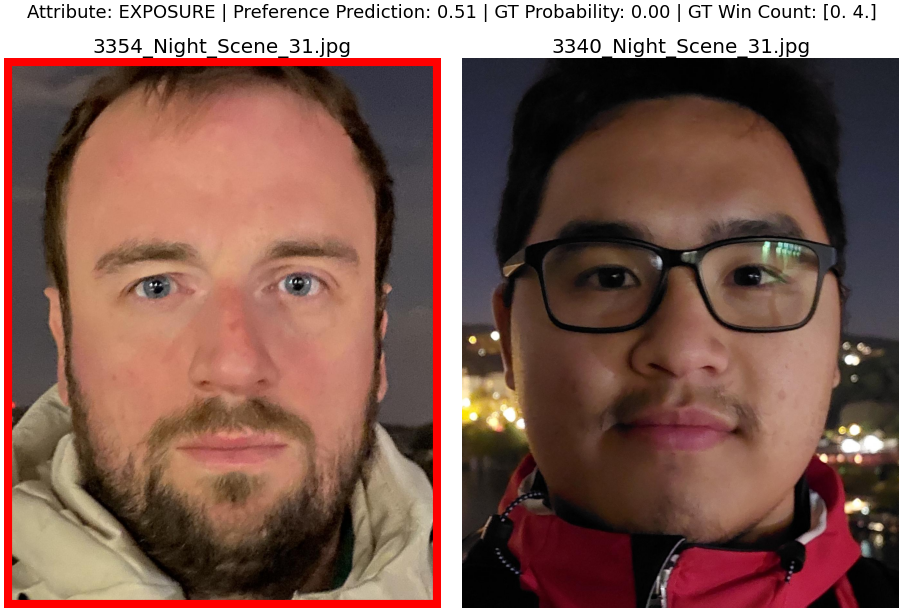} &
        \includegraphics[width=0.49\columnwidth]{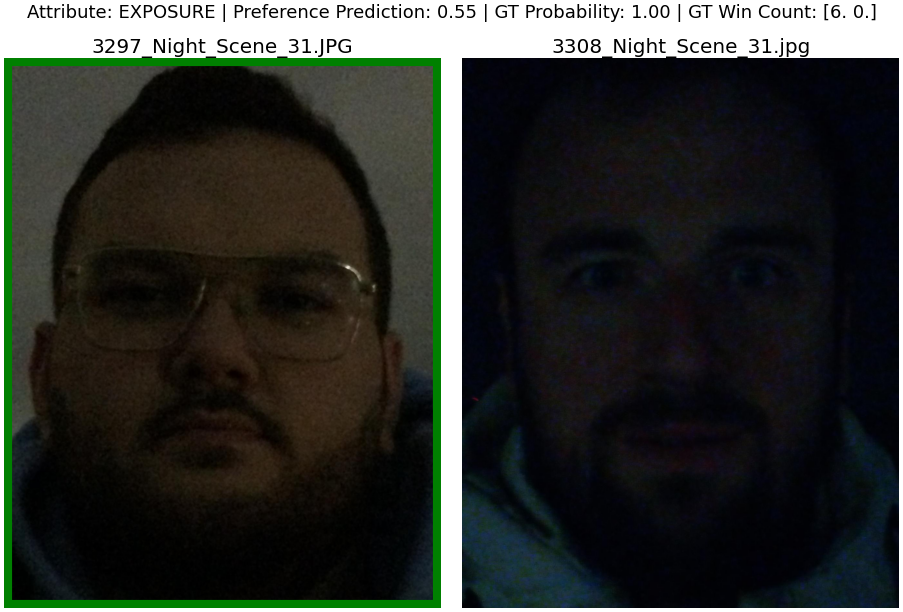} \\
    \end{tabular}
    \caption{\footnotesize{Examples of incorrect (left) and correct (right) predictions on multiple scenes for the Exposure attribute of the PIQ23 dataset. PICNIQ consistently delivers logical and precise comparisons despite the difficulty of the example. Some incorrect predictions are subject to interpretation.}}

\end{figure}

\begin{figure}[!htbp]
    \centering
    \setlength{\tabcolsep}{2pt}
    \begin{tabular}{cc}
        \includegraphics[width=0.49\columnwidth]{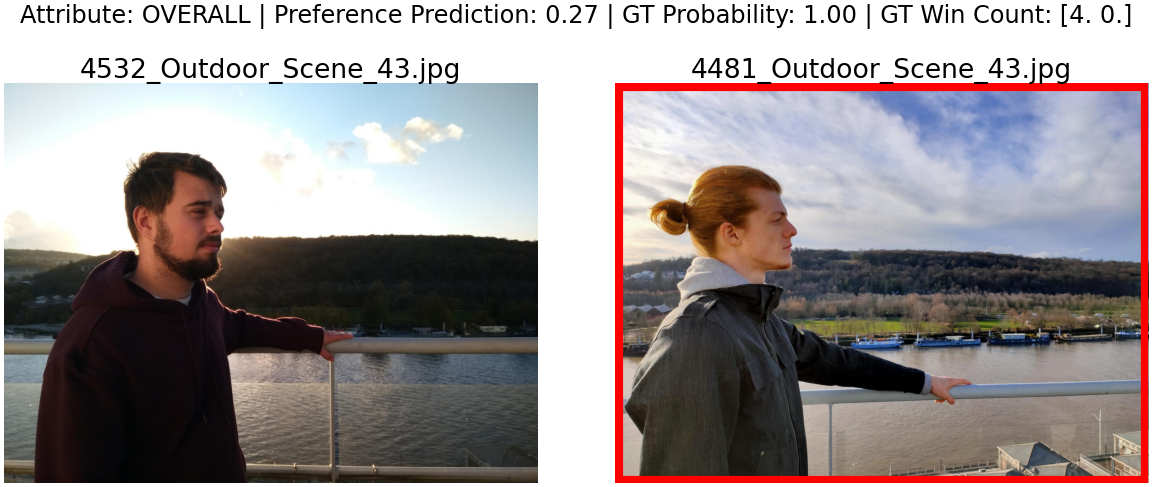} &
        \includegraphics[width=0.49\columnwidth]{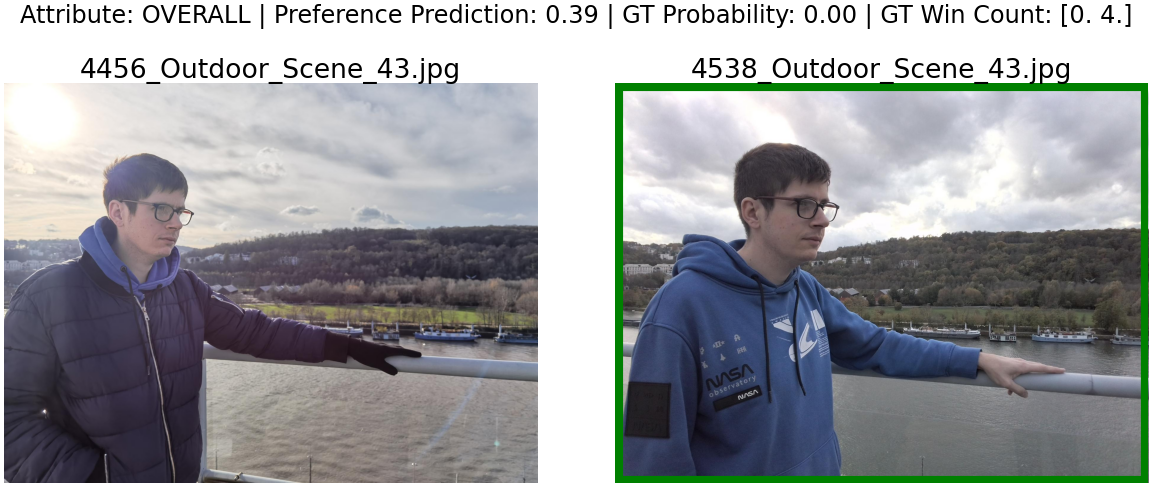} \\
        \includegraphics[width=0.49\columnwidth]{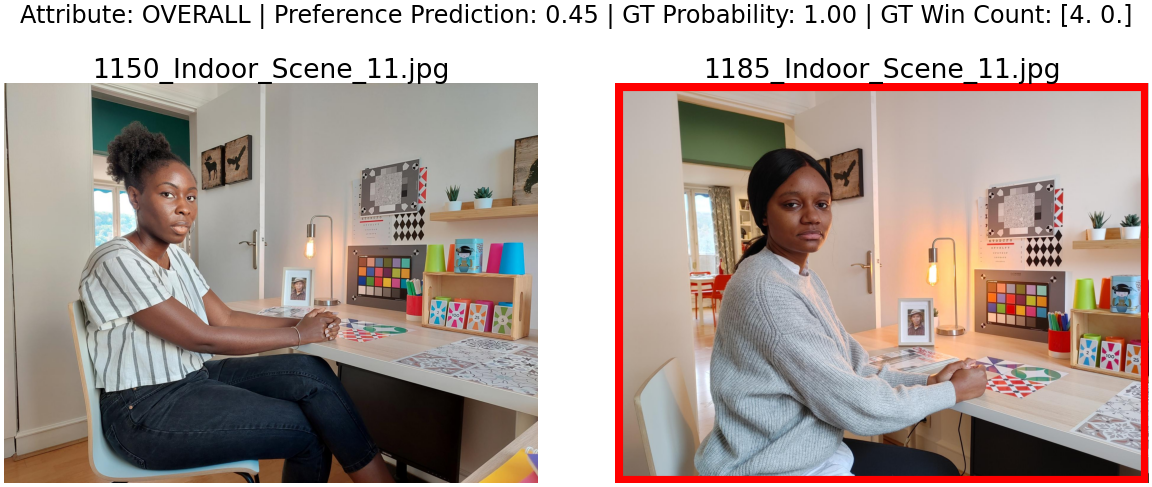} &
        \includegraphics[width=0.49\columnwidth]{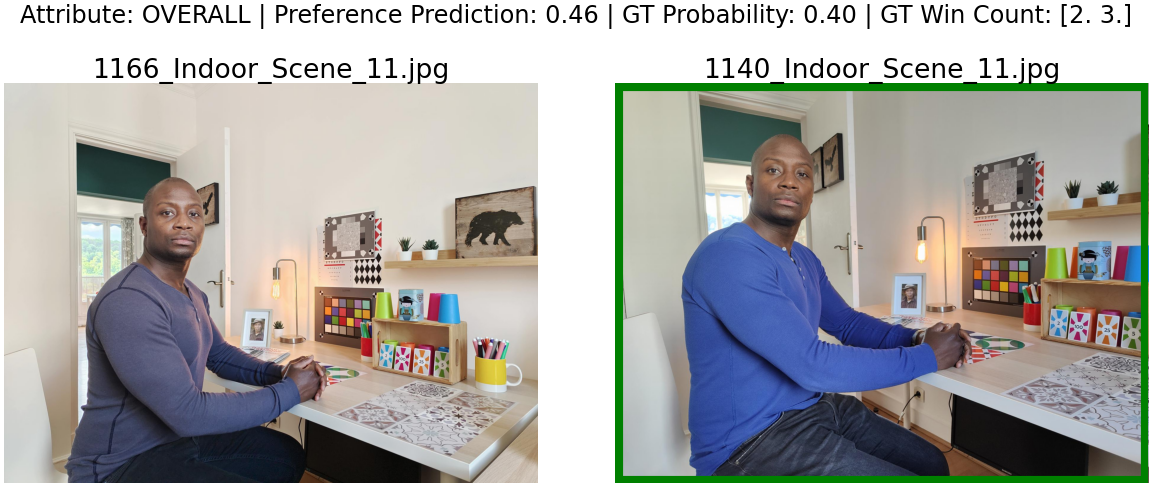} \\
        \includegraphics[width=0.49\columnwidth]{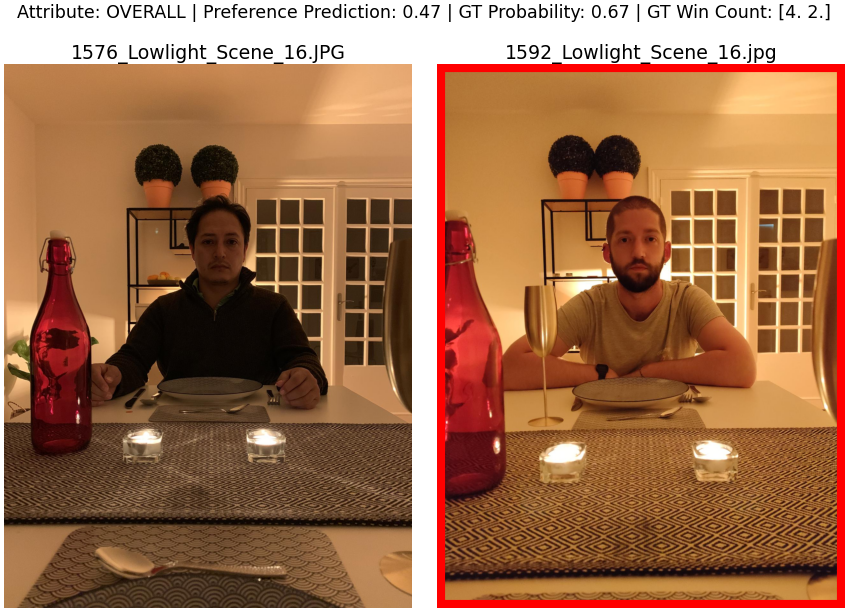} &
        \includegraphics[width=0.49\columnwidth]{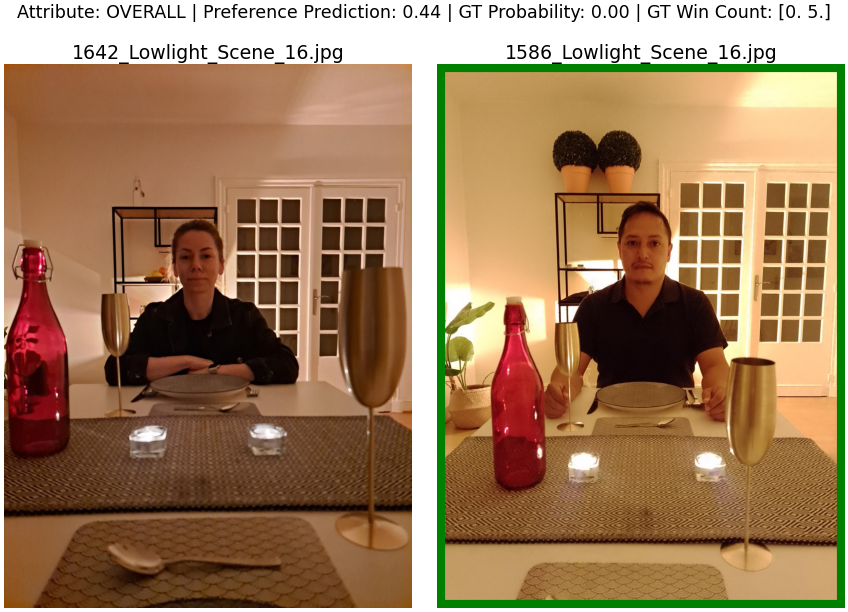} \\
        \includegraphics[width=0.49\columnwidth]{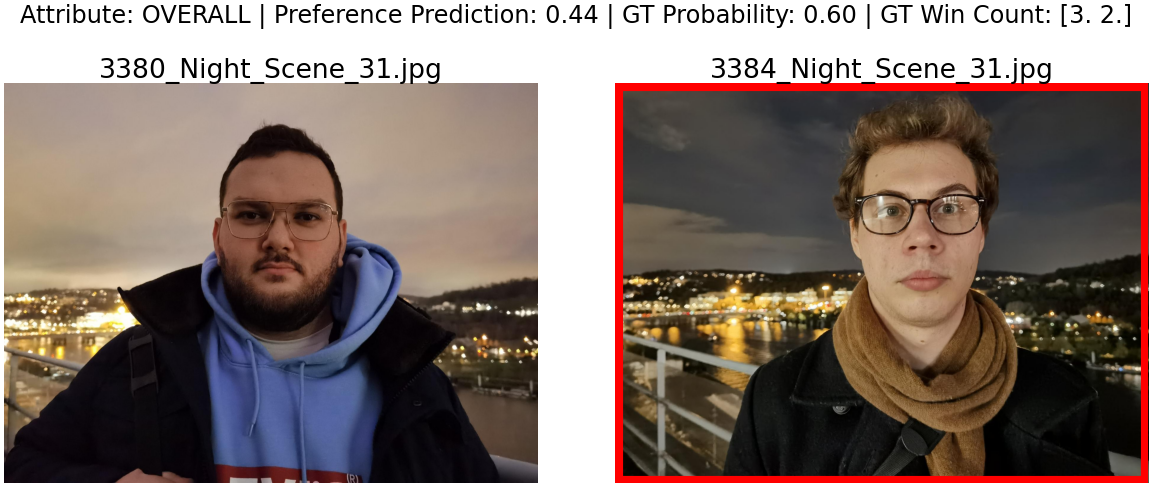} &
        \includegraphics[width=0.49\columnwidth]{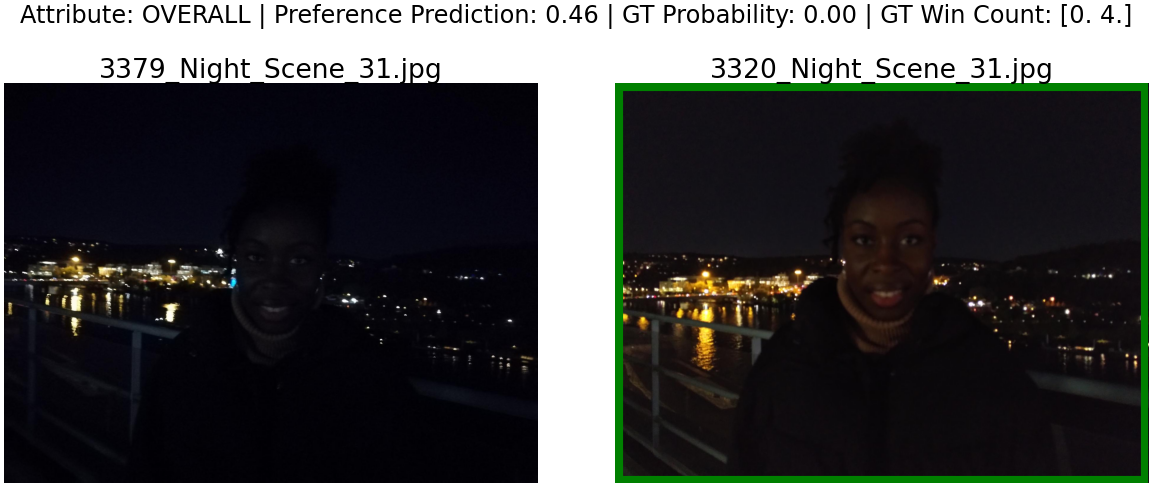} \\
        
    \end{tabular}
    \caption{\footnotesize{Examples of incorrect (left) and correct (right) predictions on multiple scenes for the Overall attribute of the PIQ23 dataset. PICNIQ consistently delivers logical and precise comparisons despite the difficulty of the example. Some incorrect predictions are subject to interpretation.}}

\end{figure}

\begin{figure}[!htbp]
    \centering
    \setlength{\tabcolsep}{2pt}
    \begin{tabular}{ccc}
        \includegraphics[width=0.32\columnwidth]{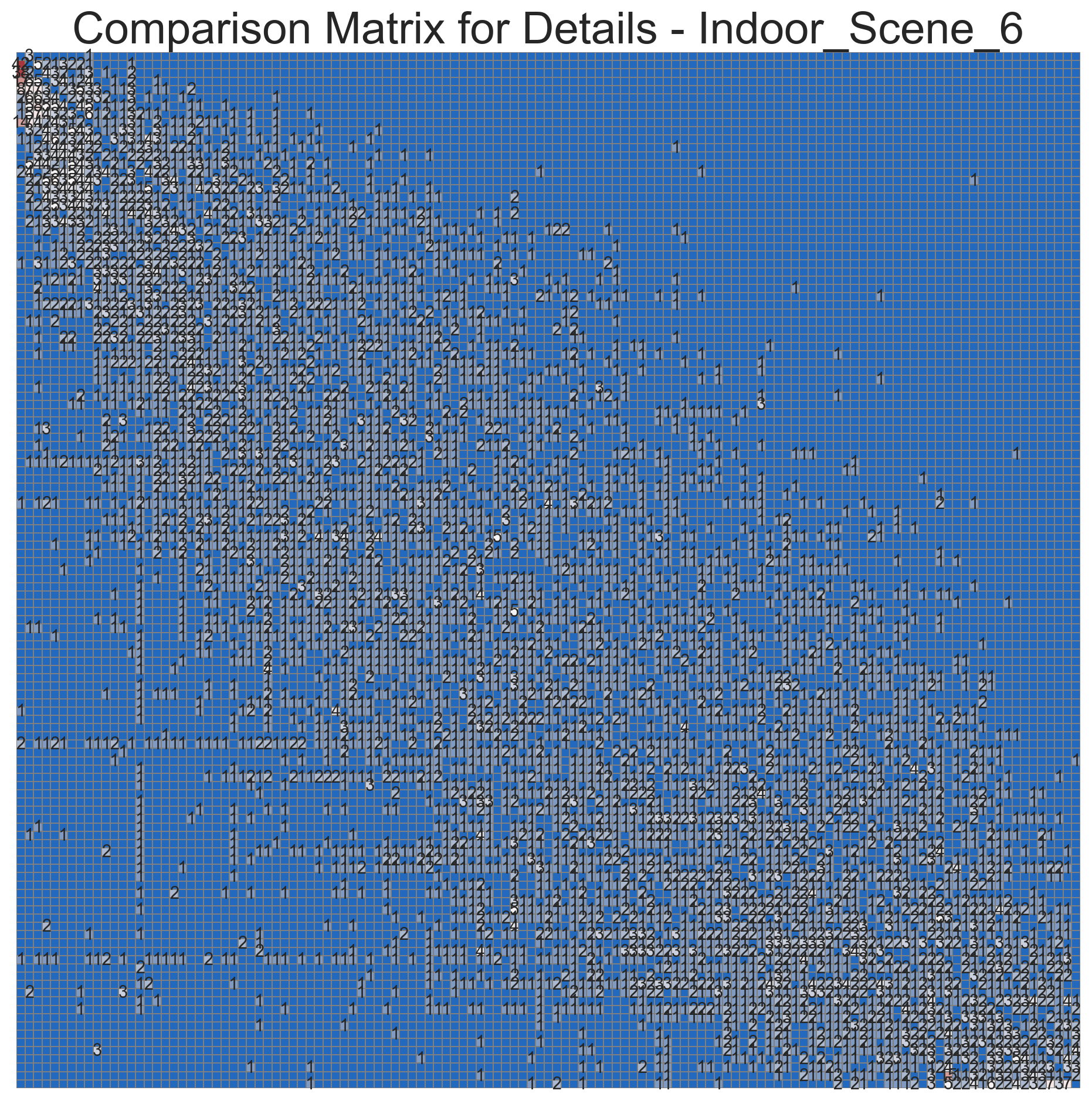} &
        \includegraphics[width=0.32\columnwidth]{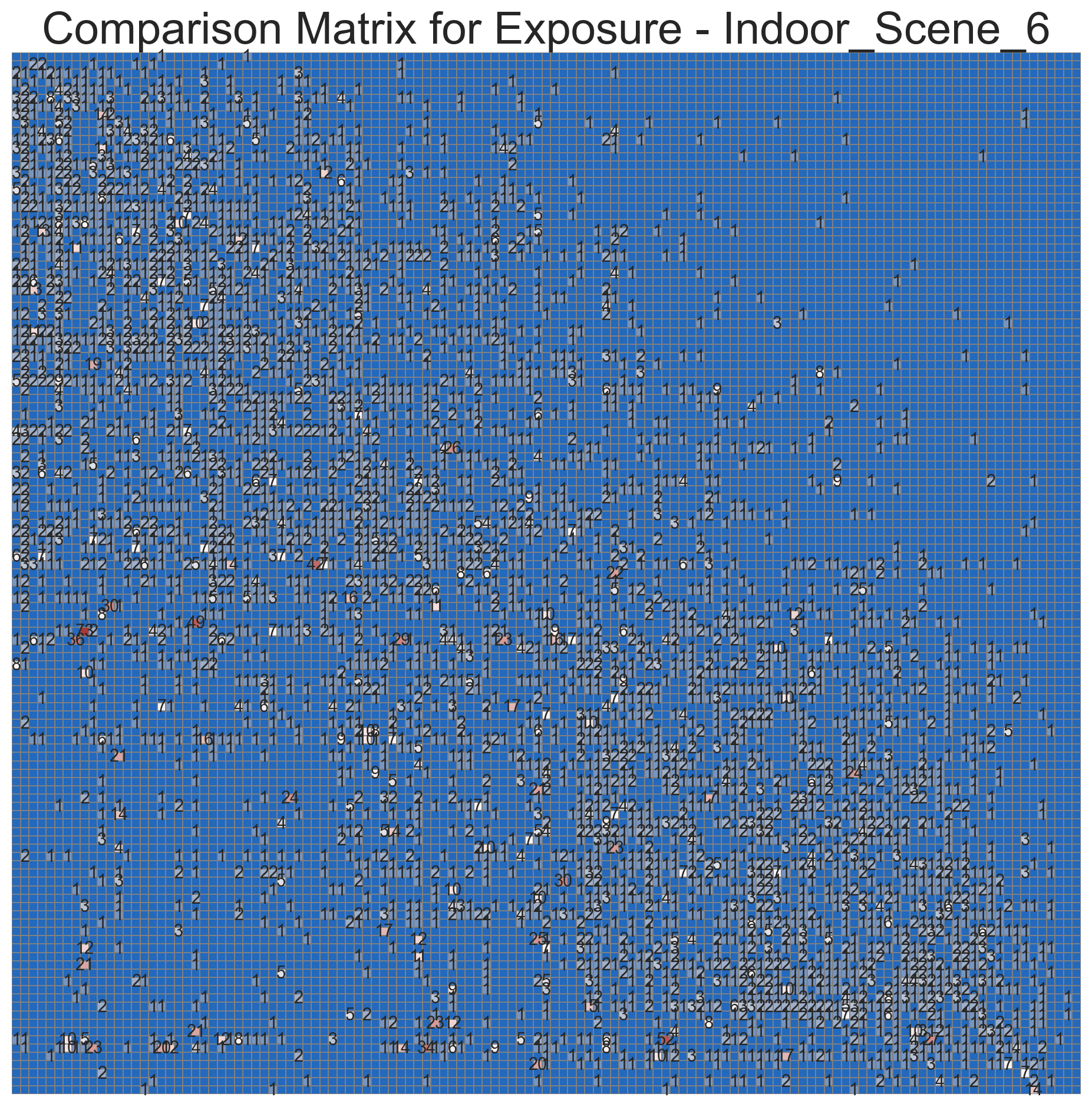} &
        \includegraphics[width=0.32\columnwidth]{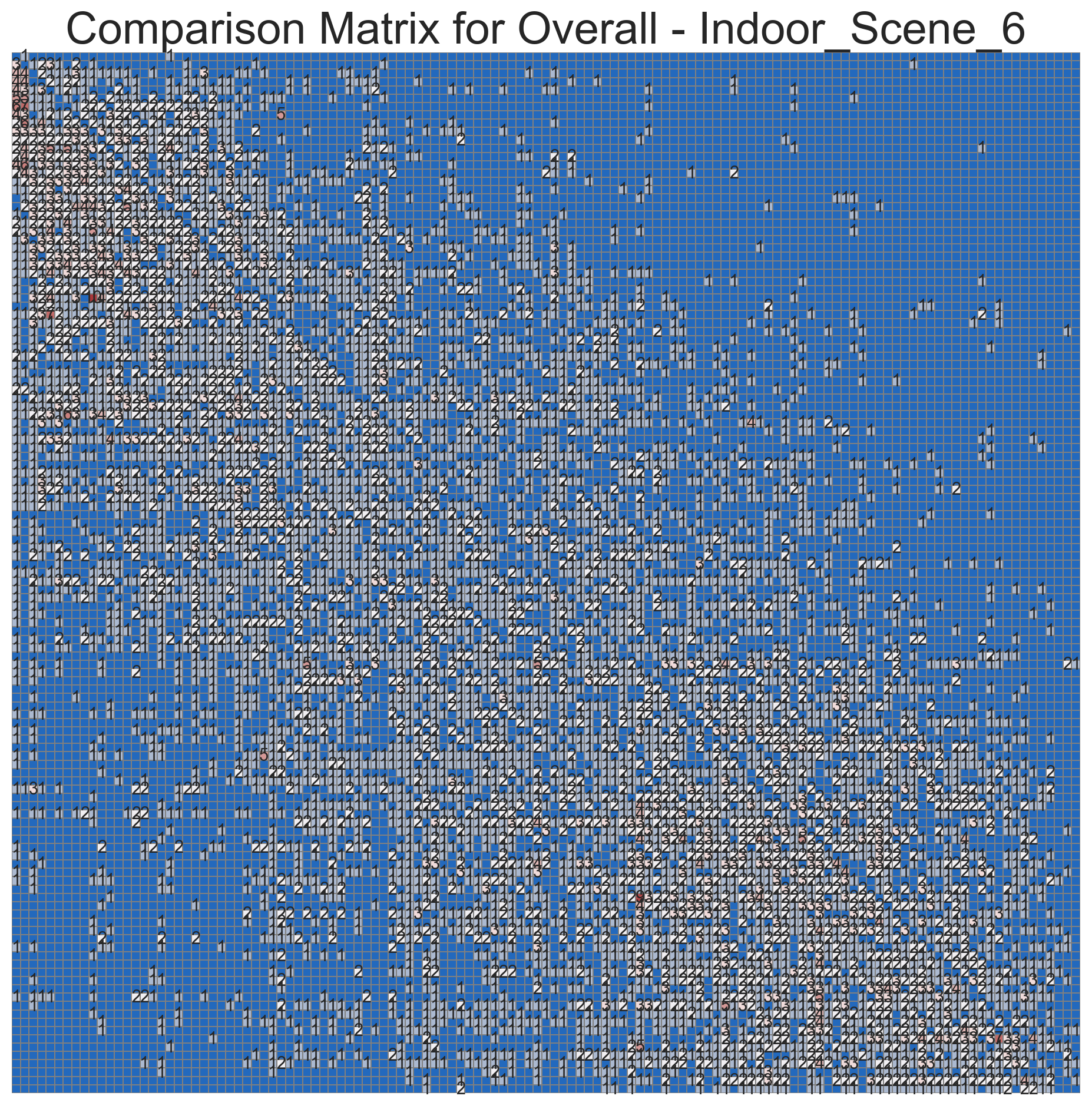} \\
        \includegraphics[width=0.32\columnwidth]{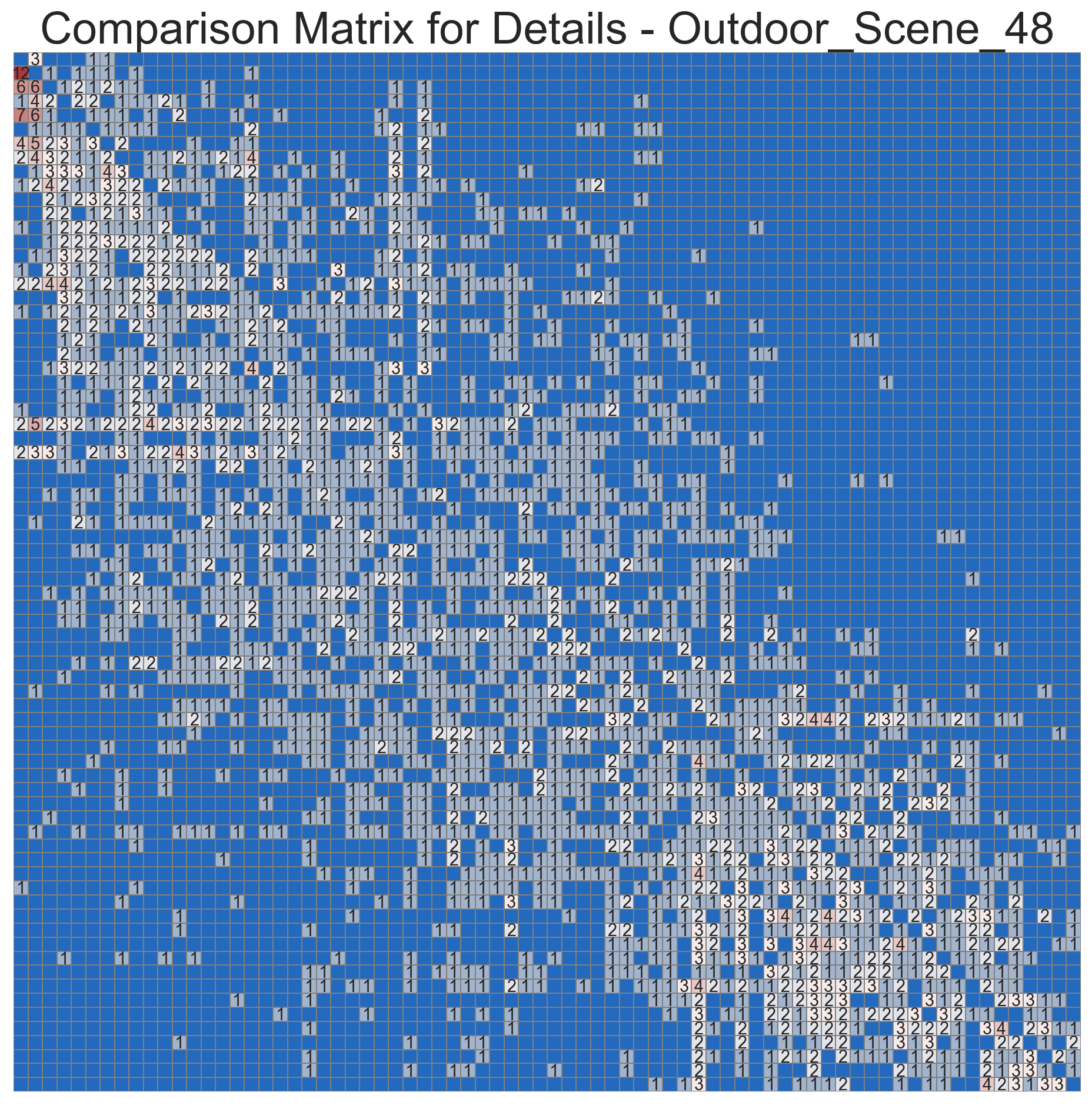} &
        \includegraphics[width=0.32\columnwidth]{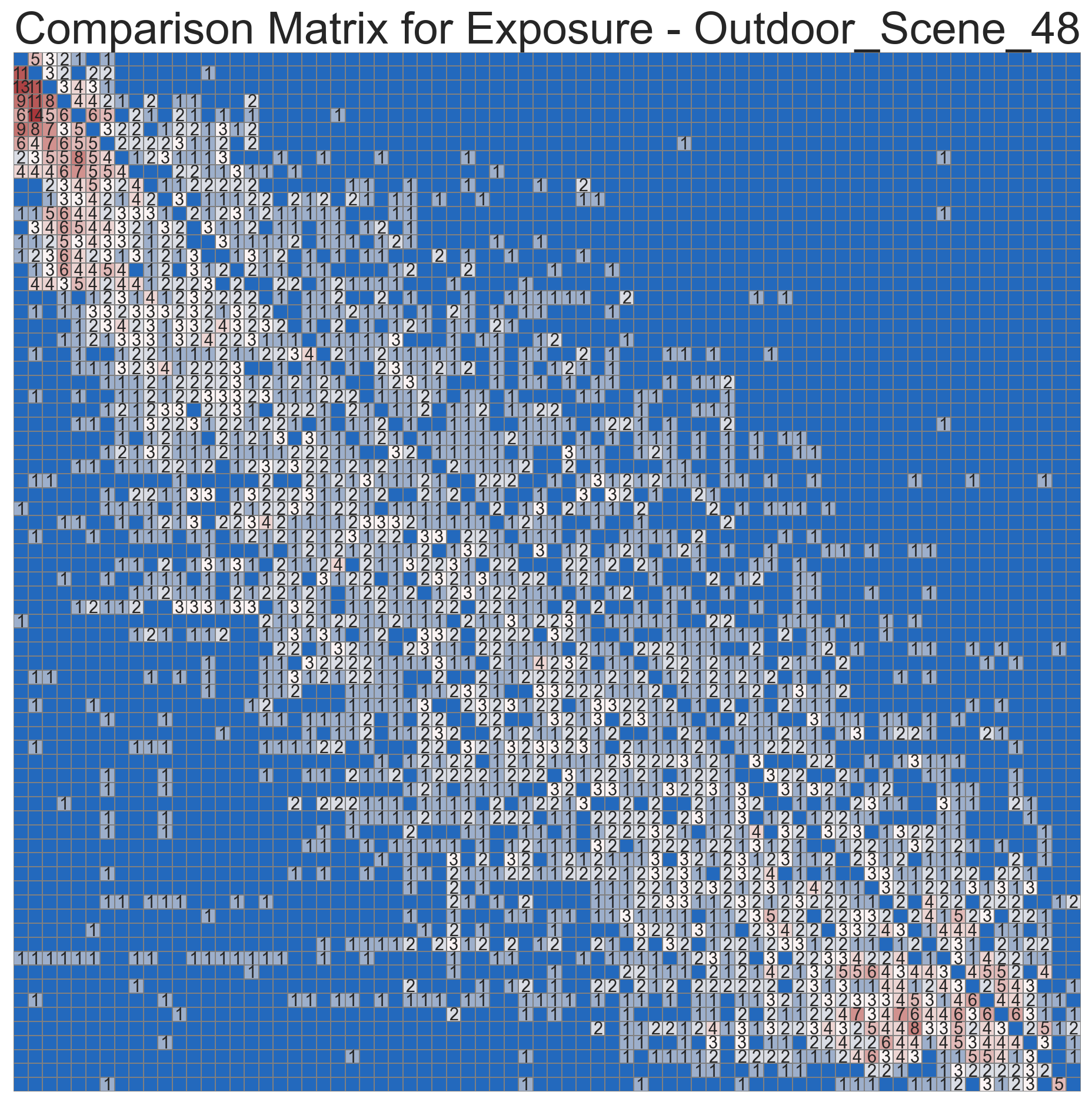} &
        \includegraphics[width=0.32\columnwidth]{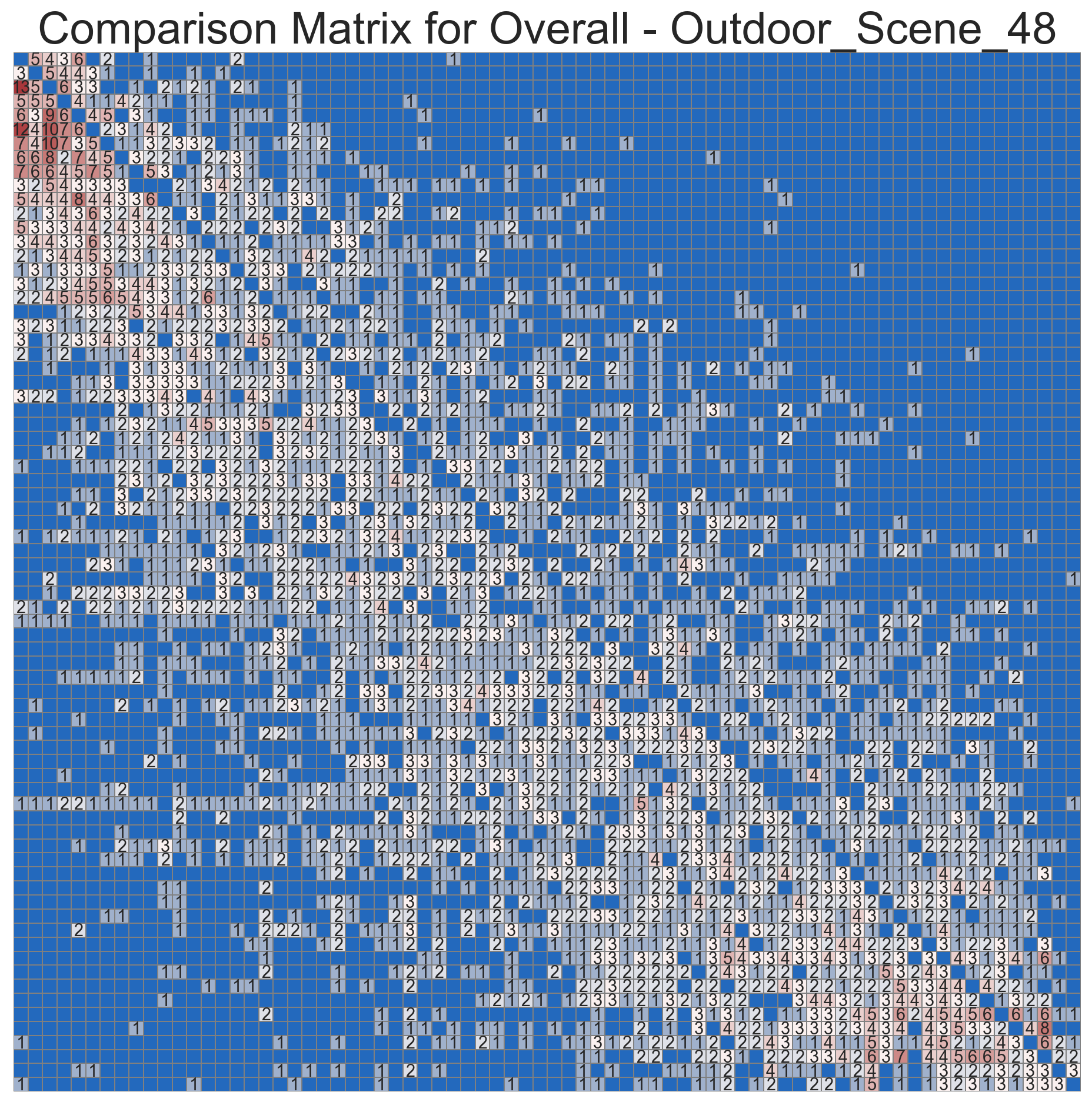} \\
        \includegraphics[width=0.32\columnwidth]{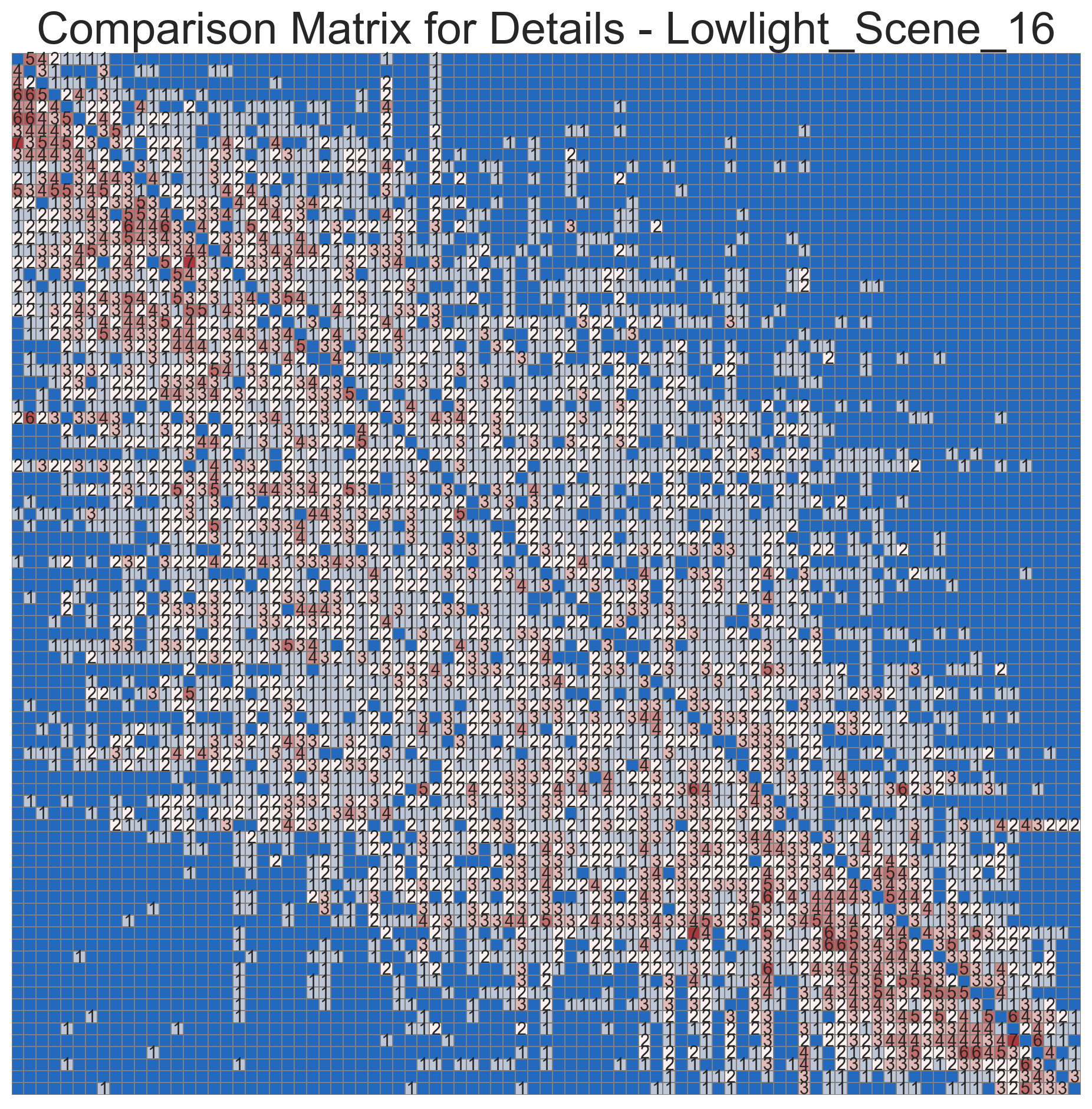} &
        \includegraphics[width=0.32\columnwidth]{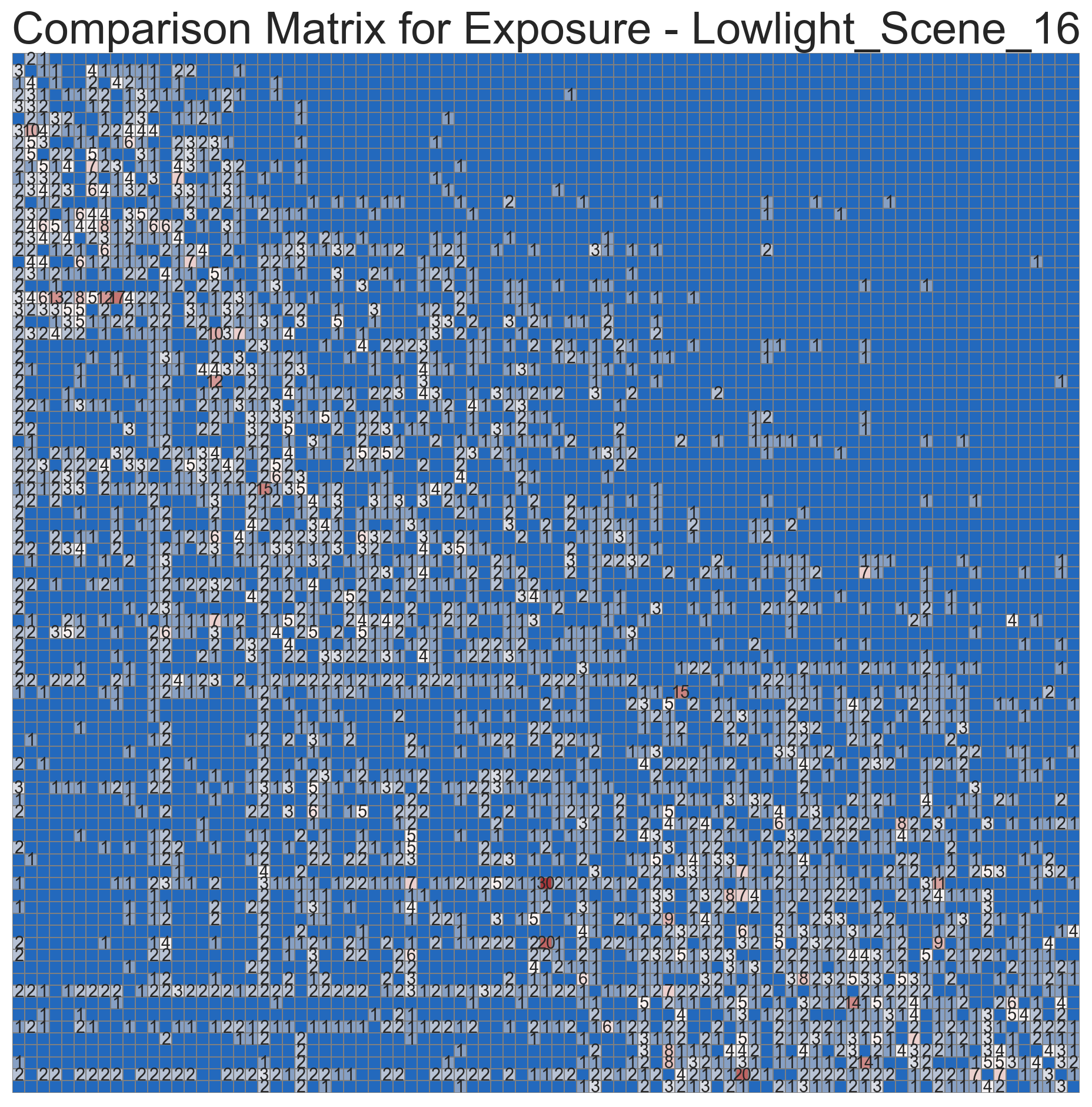} &
        \includegraphics[width=0.32\columnwidth]{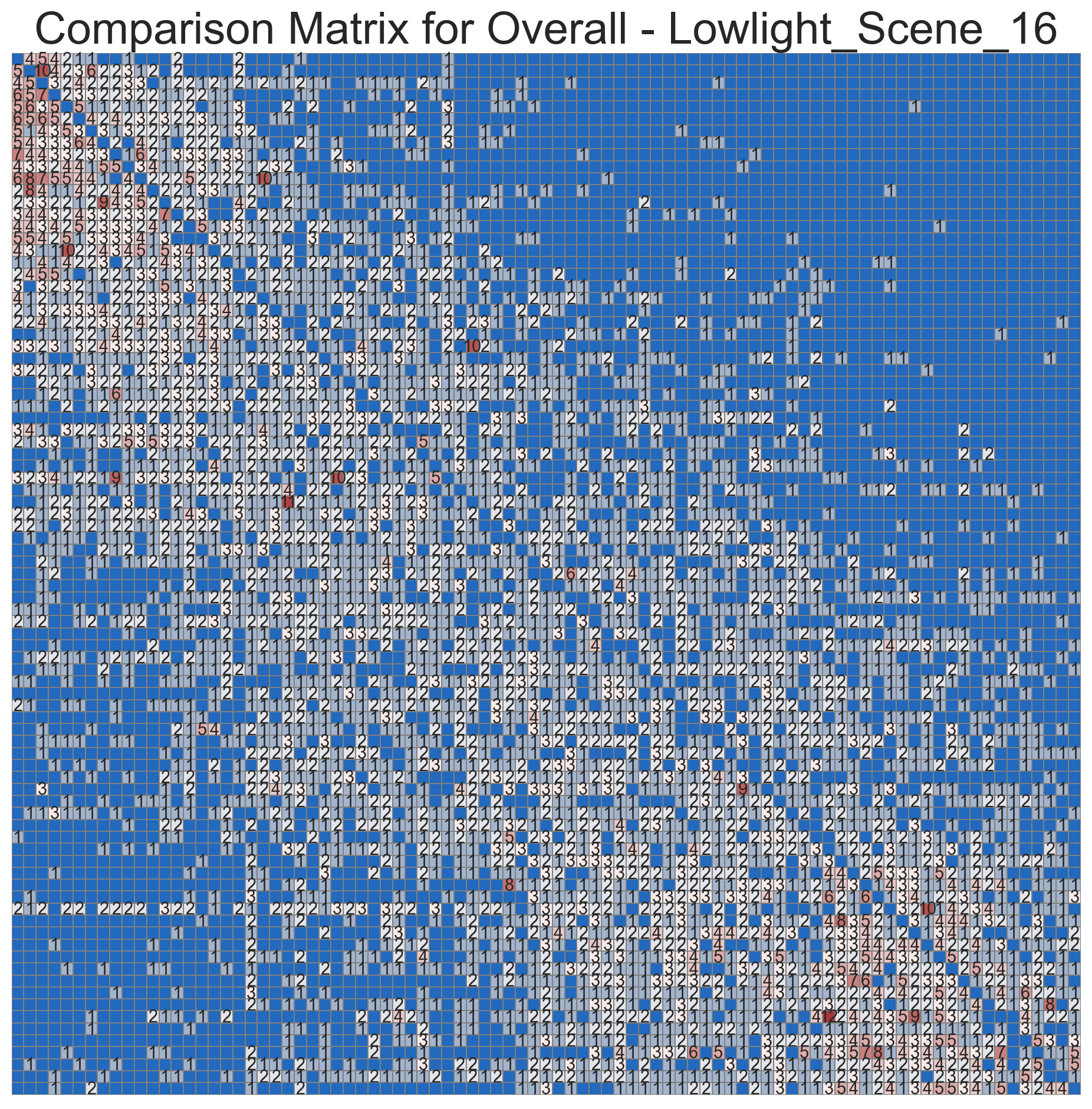} \\
        \includegraphics[width=0.32\columnwidth]{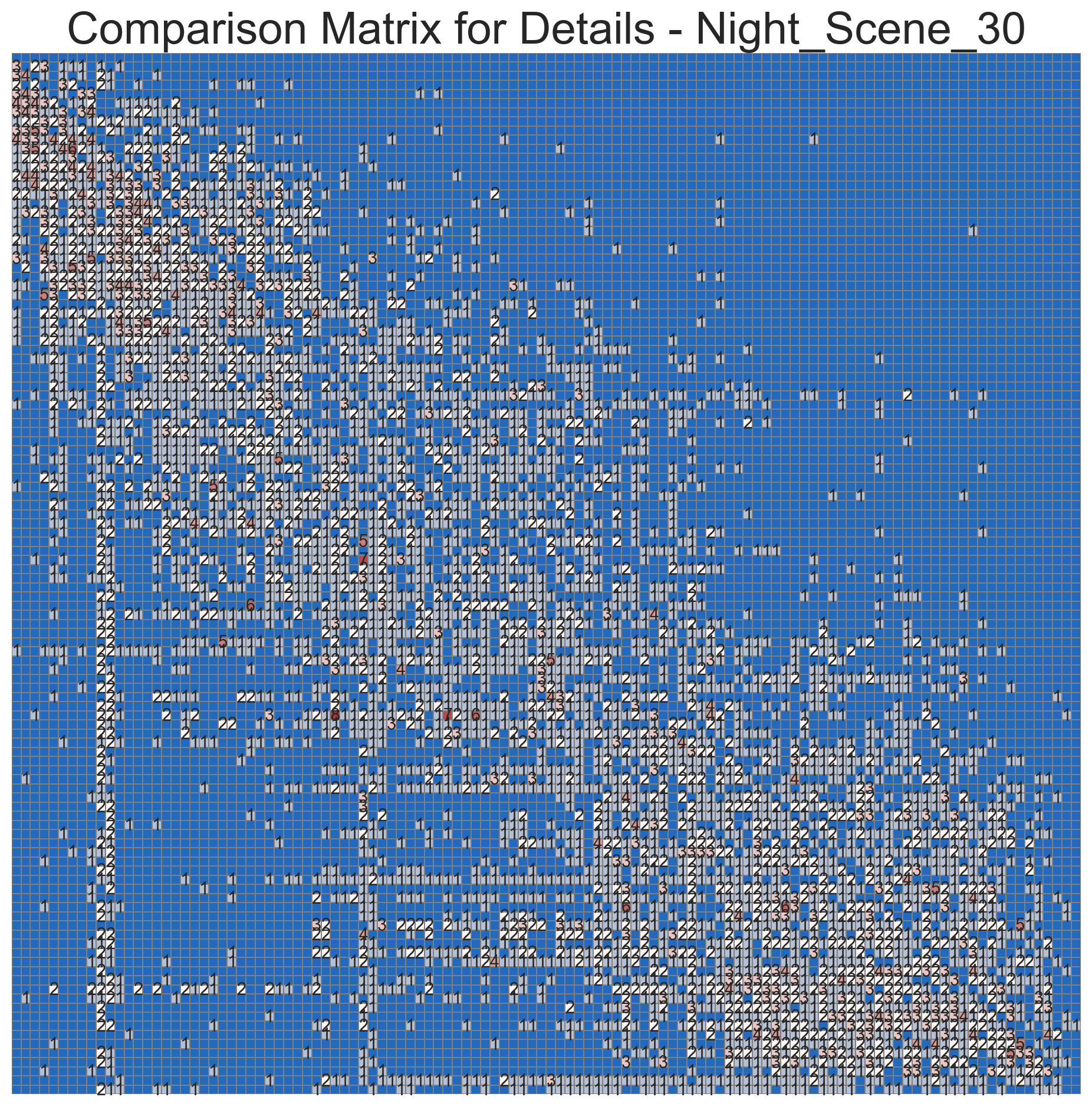} &
        \includegraphics[width=0.32\columnwidth]{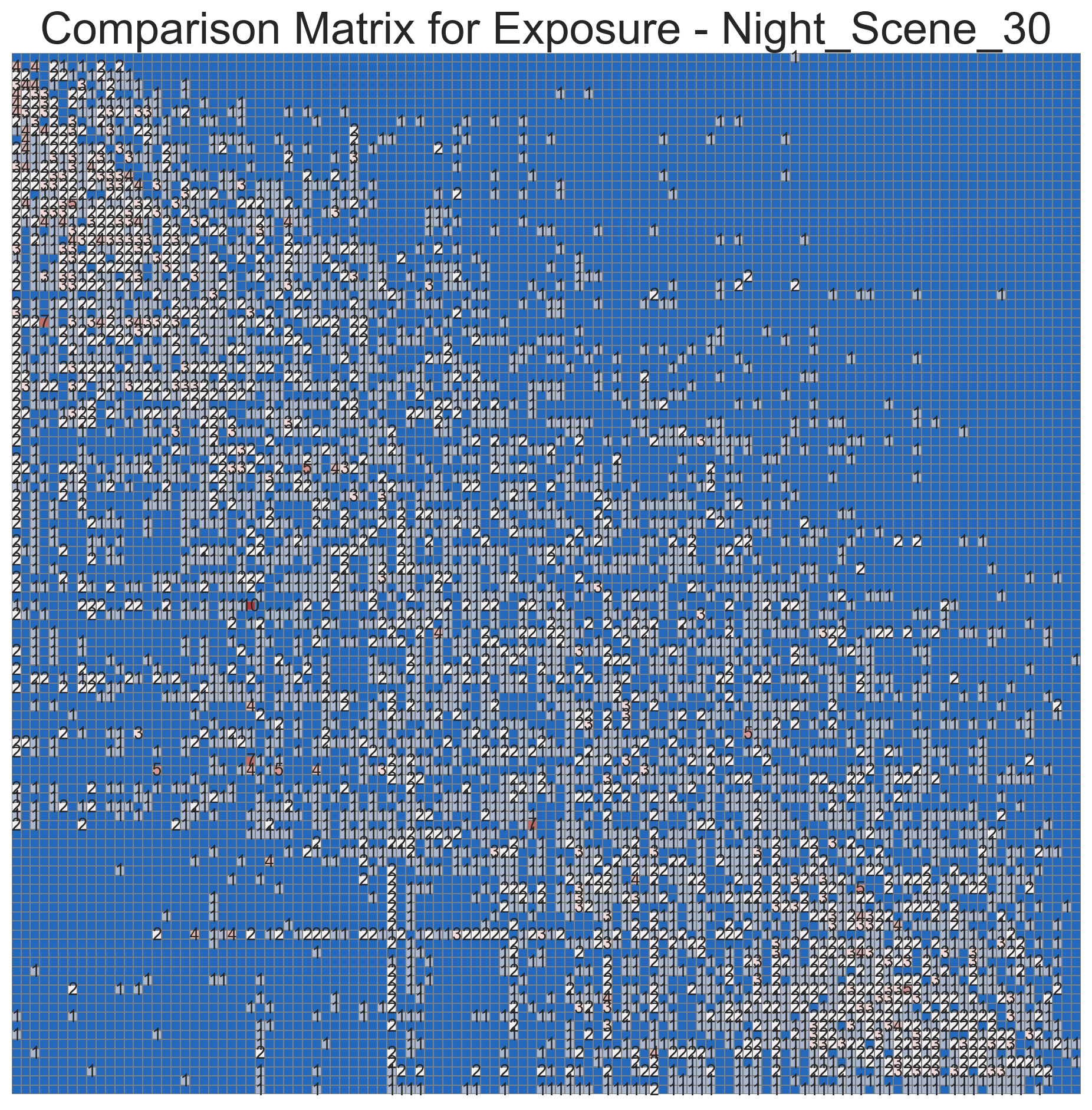} &
        \includegraphics[width=0.32\columnwidth]{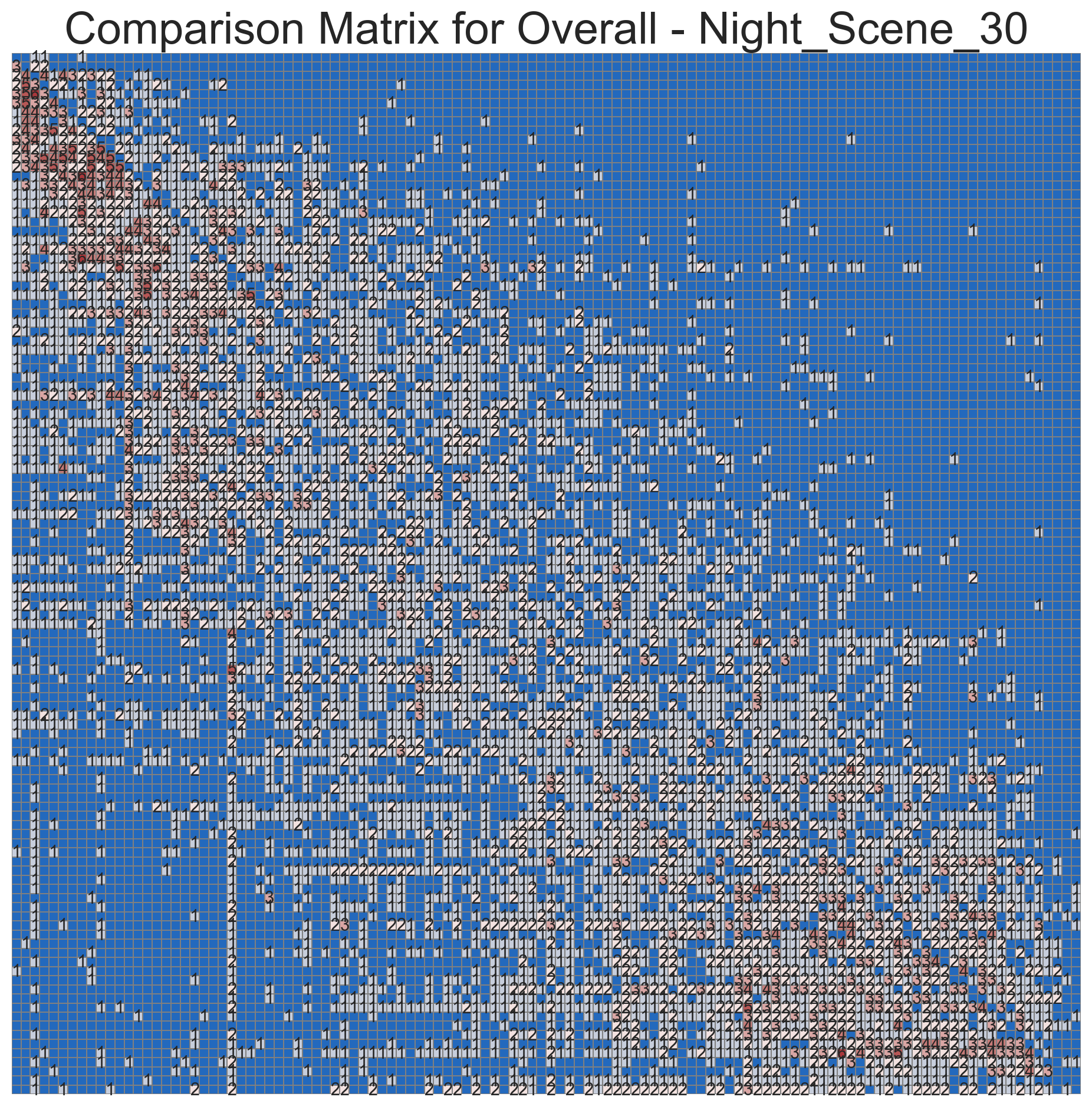} \\
    \end{tabular}
    \caption{\footnotesize{Comparison matrices for different scenes and attributes from the PIQ23 dataset. Each row represents a scene, and each column represents an attribute.}}

\end{figure}

\begin{figure}[!htbp]
    \centering
    {\includegraphics[width=0.99\columnwidth]{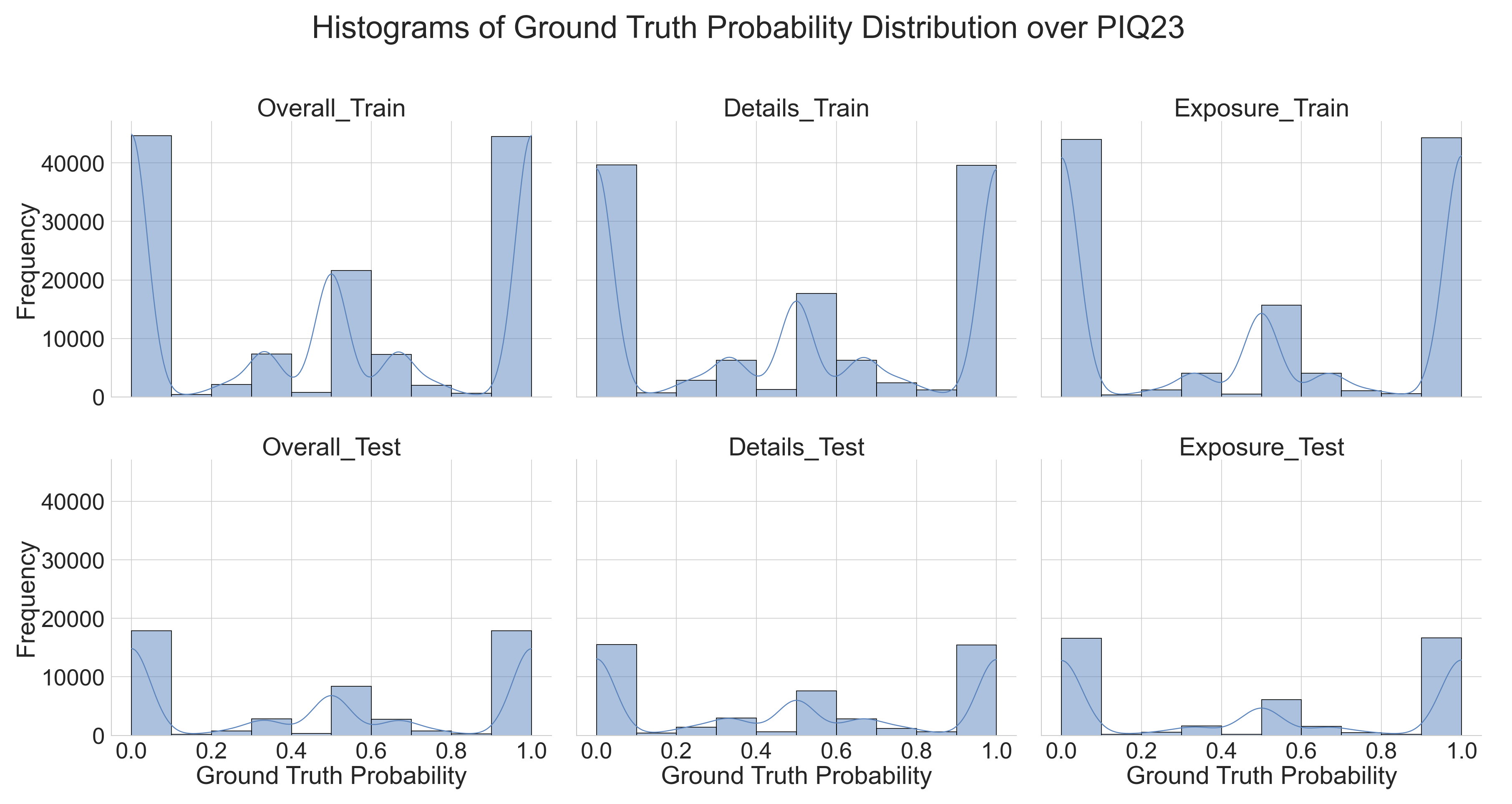}}
    \caption{Histograms of ground truths probability distribution over PIQ23 scene split over the combined scenes for all attributes. We observe that PIQ23 presents distribution imbalances towards forced-choice pairs (0s and 1s).}
\end{figure}

\begin{figure*}
    \centering
    \begin{subfigure}{0.99\textwidth}
        \centering
        \includegraphics[width=\linewidth]{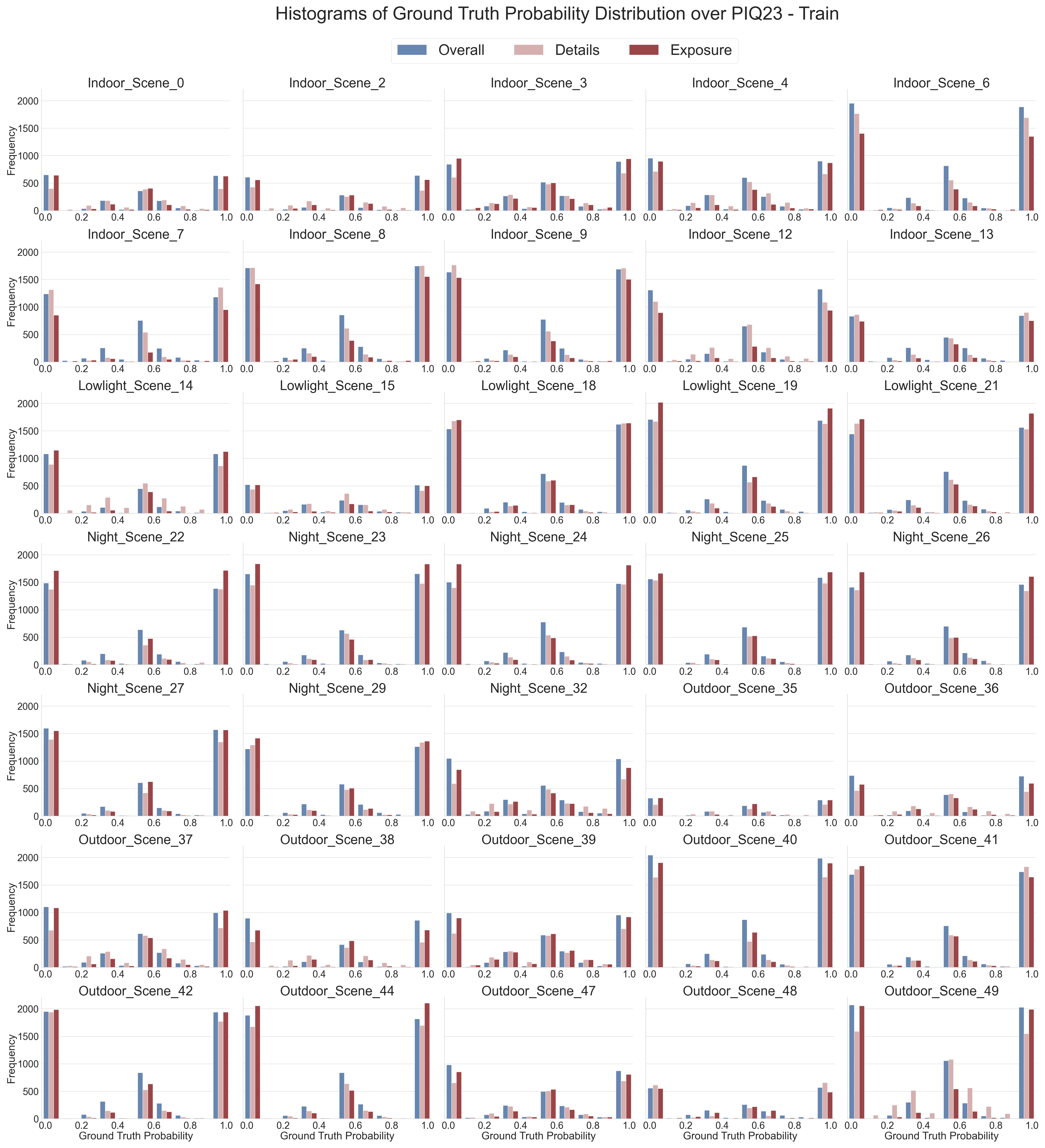}
        \caption{}\label{fig:a}
    \end{subfigure}
    \caption{(a) Histograms of ground truths probability distribution over PIQ23 for the scene split training set for all attributes.}
\end{figure*}

\begin{figure*}
    \ContinuedFloat 
    \centering
    \begin{subfigure}{0.99\textwidth}
        \centering
        \includegraphics[width=\linewidth]{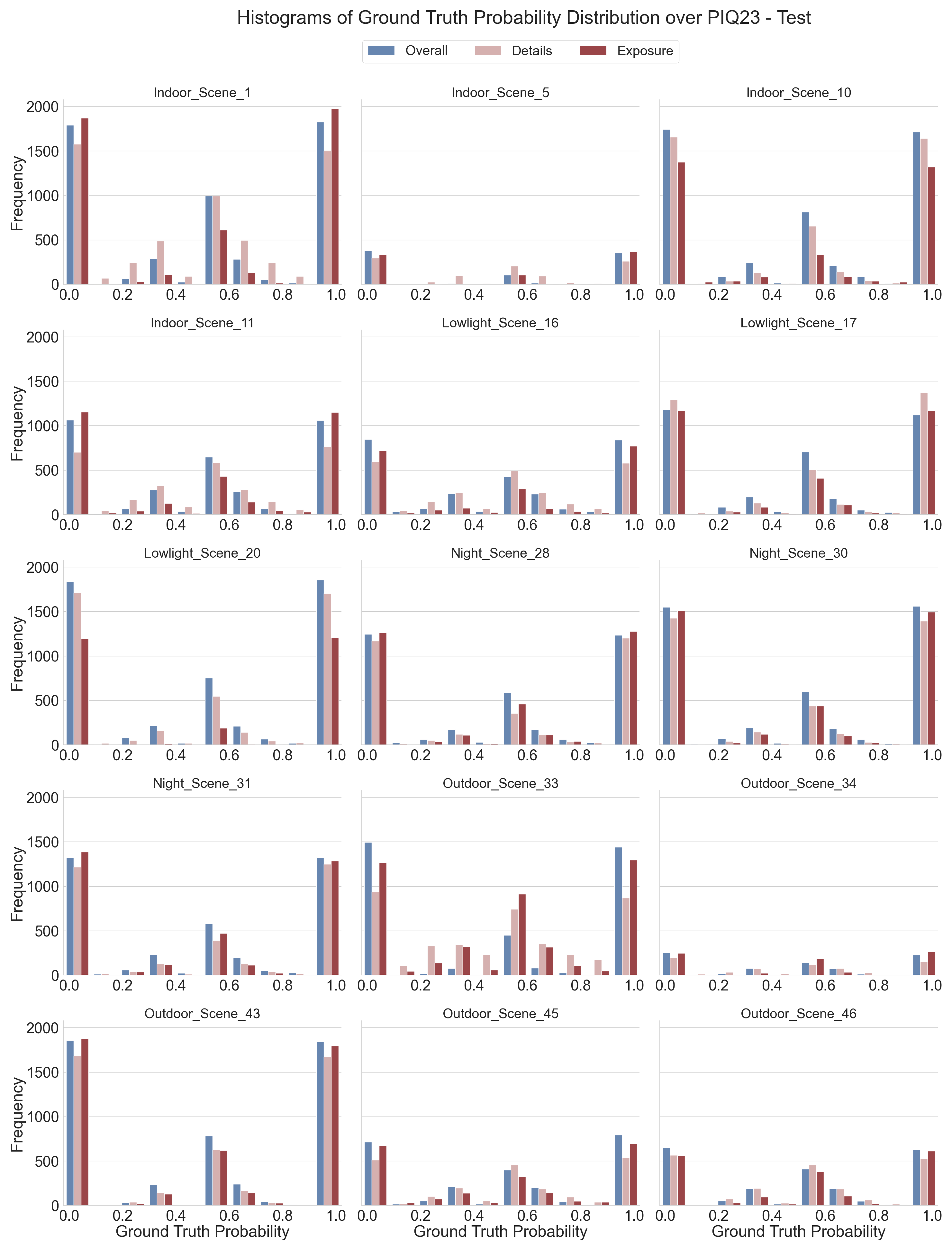}
        \caption{}\label{fig:b}
    \end{subfigure}
    \caption{(b) Histograms of ground truths probability distribution over PIQ23 for the scene split testing set for all attributes.}
\end{figure*}